\documentclass[a4paper,12pt]{report}
\usepackage{jmlr2e} 
\usepackage{smallfig}
\usepackage{times}
\usepackage{verbatim}
\usepackage{epsfig}
\RequirePackage{amssymb}
\usepackage{subfigure,amsmath,amssymb} 
\usepackage[ruled,vlined]{algorithm2e}
\usepackage{algorithmic}
\usepackage{appendix}
\usepackage{bbm}
\bibpunct{[}{]}{;}{a}{}{,}

\long\def\ifundefined#1#2#3{\expandafter\ifx\csname#1\endcsname\relax
#2\else#3\fi}

\ifundefined{DEFNSTYPRESENT}{\def\thmcolon{\hspace{-.85em} {\bf
:} }}{\def\thmcolon{\relax}} 

\newtheorem{THEOREM}{Theorem}[section]
\newcommand{\thm}{\begin{THEOREM} \thmcolon \rm}
\newcommand{\thmwname}[1]{\begin{THEOREM} {\bf {#1}} \thmcolon \rm}
\newcommand{\lem}{\begin{lemma} \rm}
\newcommand{\pro}{\begin{proposition}}
\newcommand{\dfn}{\begin{definition}}
\newcommand{\rem}{\begin{remark}}
\newcommand{\xam}{\begin{example}}
\newcommand{\cor}{\begin{corollary}}
\newcommand{\prf}{\noindent{\bf Proof:} }
\newcommand{\ethm}{\end{THEOREM}}
\newcommand{\elem}{\end{lemma}}
\newcommand{\epro}{\end{proposition}}
\newcommand{\edfn}{\bbox\end{definition}}
\newcommand{\erem}{\bbox\end{remark}}
\newcommand{\exam}{\bbox\end{example}}
\newcommand{\ecor}{\end{corollary}}
\newcommand{\eprf}{\bbox\vspace{0.1in}}
\newcommand{\beqn}{\begin{equation}}
\newcommand{\eeqn}{\end{equation}}
\newcommand{\beqa}{\begin{eqnarray*}}
\newcommand{\eeqa}{\end{eqnarray*}}

\newcommand{\denselist}{\itemsep 0pt\partopsep 0pt}
\newcommand{\bitem}{\begin{itemize}\denselist}
\newcommand{\eitem}{\end{itemize}}
\newcommand{\benum}{\begin{enumerate}\denselist}
\newcommand{\eenum}{\end{enumerate}}

\newcommand{\commentout}[1]{}

\newcommand{\bbox}{\vrule height7pt width4pt depth1pt}

\newcommand{\thmref}[1]{Theorem~\ref{#1}}

\newcommand{\tabref}[1]{Table~\ref{#1}}
\newcommand{\figref}[1]{Figure~\ref{#1}}
\newcommand{\secref}[1]{Section~\ref{#1}}

\newcommand{\chapref}[1]{Chapter~\ref{#1}}
\newcommand{\proref}[1]{Proposition~\ref{#1}}
\newcommand{\lemref}[1]{Lemma~\ref{#1}}

\newcommand{\algref}[1]{Algorithm~\ref{#1}}

\reversemarginpar

\newcommand{\points}[1]{}

\newcommand{\answer}[1]{}

\newcommand{\ssq}{{\sigma^2}}
\newcommand{\stp}{\sqrt{2\pi}}
\newcommand{\lrh}{\ell_{hinge}^{rob}}
\newcommand{\erf}{\text{erf}}
\newcommand{\inverf}{{\erf_{{\mbox\tiny inv}}}}

\newcommand{\needcite}[1]{}

\newcommand{\be}{\begin{equation}}
\newcommand{\ee}{\end{equation}}
\newcommand{\bea}{\begin{eqnarray*}}
\newcommand{\eea}{\end{eqnarray*}}

\newcommand{\lag}{\mathcal{L}}
\newcommand{\ww}{\boldsymbol{w}} 
\newcommand{\xx}{\boldsymbol{x}} 
\newcommand{\nn}{\boldsymbol{n}} 
\newcommand{\uu}{\boldsymbol{u}} 
 
\newcommand{\zz}{\boldsymbol{z}} 
\newcommand{\alv}{\boldsymbol{\alpha}} 
\newcommand{\deltav}{\boldsymbol{\delta}} 

\newcommand{\vv}{\boldsymbol{v}}

\newcommand{\WW}{\boldsymbol{W}}

\newcommand{\ignore}[1]{}

\renewcommand{\eqref}[1]{Equation \ref{#1}}
\newcommand{\appref}[1]{Appendix \ref{#1}}

\newcommand{\ri}{\right}
\newcommand{\lf}{\left}

\SetKwComment{comm}{//}{}

\graphicspath{{figs/}}

\begin{document}

\titlepage
\thispagestyle{empty} \hspace{1 cm} \vspace{ 2 cm}
\begin{center}
\Huge Gaussian Robust Classification
\vspace{3 cm}

\large A thesis submitted in partial fulfillment of the
\\requirements for the degree of Master of Science
\\
\vspace{1 cm} by
\\
\LARGE Ido Ginodi

\vspace{2 cm} \large Supervised by Dr. Amir Globerson
\\
\vspace{2 cm} December 2010
\\
\vspace{1 cm} \normalsize The School of Computer Science and Engineering\\
The Hebrew University of Jerusalem, Israel
\end{center}

\newpage
\begin{abstract}
\emph{Supervised learning} is all about the ability to generalize knowledge. Specifically, the goal of the learning is to train a classifier using training data, in such a way that it will be capable of classifying new unseen data correctly. In order to acheive this goal, it is important to carefully design the learner, so it will not \emph{overfit} the training data. The later can be done in a couple of ways, where adding a regularization term is probably the most common one. The statistical learning theory explains the success of the regularization method by claiming that it restricts the complexity of the learned model. This explanation, however, is rather abstract and does not have a geometric intuition.

The generalization error of a classifier may be thought of as correlated with its robustness to perturbations of the data. Namely, if a classifier is capable of coping with distrubance, it is expected to generalize well. Indeed, it was established that the ordinary SVM formulation is equivalent to a robust formulation, in which an adversary may displace the training and testing points within a ball of pre-determined radius (\cite{XuCaMa09}).

In this work we explore a different kind of robustness. We suggest changing each data point with a Gaussian cloud centered at the original point. The loss is evaluated as the expectation of an underlying loss function on the cloud. This setup fits the fact that in many applications, the data is sampled along with noise. We develop a robust optimization (RO) framework, in which the adversary chooses the covariance of the noise. In our algorithm named GURU, the tuning parameter is the variance of the noise that contaminates the data, and so it can be estimated using physical or applicative considerations. Our experiments show that this framework generates classifiers that perform as well as SVM and even slightly better in some cases. Generalizations for Mercer kernels and for the multiclass case are presented as well. We also show that our framework may be further generalized, using the technique of \emph{convex perspective} functions.

\end{abstract}

\tableofcontents
\newpage
\chapter{Introduction}
\label{section-introduction}

\section{Motivation}
The ability to understand new unseen data, based on knowledge that was gained using a training sample, is probably the main goal of machine learning. In the supervised learning setup, one is given a training set, consists of data samples along with labels indicating their 'type' or 'class'. The learning task in this case is to develop a decision rule, which will allow predicting the correct label of unfamiliar data. 

As the main goal is to be able to generalize, it makes sense to design the learning process so it reflects the conditions under which the classifier is going to be tested and used. In many real world applications, the data we are given is corrupted by noise. The noise may be either inherent to the process that generates the data or adversarial. Examples to an inherent noise include a noisy sensor and natural variability of the data. Adversarial noise is present for example in spam emails. In either way, it is vital to learn how to classify when it is present. We suggest to do it by preparing for the worst case. Amongst all noise distribution that have a bounded power (i.e. bounded covariance), the Gaussian noise is believed to be the most problematic, since it has the maximal entropy.

By designing a classifier that is robust to Gaussian noise, we are able to learn and generalize well, without the need to introduce an explicit regularization term. In that respect, our work aims at shading more light on the connection between robustness and generalization. 

\section{The supervised learning framework}

Formally speaking, the supervised learning setup consists of three major components:
\begin{enumerate}
\item \textbf{Data.} We denote $\mathcal{X}$ the sample space, in which the data samples live (i.e. the objects one tries to classify. e.g., vector representation of handwritten digits). Alongside the sample space, we are given the label set, denoted $\mathcal{Y}$. This set contains the various classes to which the data points may be assigned (e.g., $0,1,\ldots,9$ in the handwritten digits example). A distribution $\mathcal{D}$ is defined over $\mathcal{X}\times\mathcal{Y}$, and dictates the probability to sample a data point $\xx\in\mathcal{X}$ along with a label $y\in\mathcal{Y}$.
In our discussion we will restrict ourselves to the Euclidean case, namely $\mathcal{X}=\mathbb{R}^d$. Unless stated otherwise, we assume a binary setting, in which $\mathcal{Y}=\{+1,-1\}$.
\item \textbf{Hypothesis class.} In the learning process, one considers candidate hypotheses taken out of the class $\mathcal{H}$. This class consists of functions from $\mathcal{X}$ to $\mathcal{Y}$. Its contents reflect some kind of prior data about the problem at hand. A well known example is the class of half-spaces, defined as
\be
\mathcal{H}_{half-space}=\left\{
	\phi_{\ww}(\xx)=\text{sgn}(\ww^T\xx) 	\big|\phi_{\ww}:\mathbb{R}^d\to\{+1,-1\},\;\ww\in\mathbb{R}^d\right\}
\ee
\item \textbf{Loss measure.} The means to measure the performance of a specific instance $h\in\mathcal{H}$ is the loss function, $\ell:\mathcal{X}\times\mathcal{Y}\times\mathcal{H}\to\mathbb{R}_+$
The most intuitive loss function in the binary case is the zero-one loss, defined by
\be
\ell_{0-1}(\xx,y;h)=\mathbbm{1}_{[h(\xx)\ne y]}
\ee
\end{enumerate}

The learning task is to find the classifier $h^*\in\mathcal{H}$ which is optimal, in the sense that it minimizes the \emph{actual risk}, defined as
\be
\text{err}(h)=\mathbb{E}_{(\xx,y)\sim\mathcal{D}}\ell(\xx,y;h)
\ee
Most of the times, however, it is the case that $\mathcal{D}$ is unknown. Even in the rare cases in which it is known, it is not always possible to optimize the expectation over it. The learner is thus given a \emph{training set} $\mathcal{S}\subseteq\mathcal{X}\times\mathcal{Y}$ of i.i.d. samples. The learning task in that case is to minimize the \emph{empirical risk}, defined as
\be
\hat{\text{err}}(h)=\frac{1}{M}\sum_{m=1}^M{\ell(\xx^m,y^m;h)}
\ee
where $\mathcal{S}=\left\{(\xx^m,y^m)\right\}_{m=1}^M$. This technique is called \emph{empirical risk minimization} (ERM).
It is important to keep in mind that although the technical tool is ERM, the objective is always to have the actual risk as low as possible.

Sometimes, however, this is not the case. That is, in spite of the fact that the learned decision rule is capable of classifying the training data, it fails to do so on fresh test data. In this case we say that the \emph{generalization error} is high, although the training error is low. The reason for such a failure is most often \emph{overfitting}. In this situation, the learned classifier fits the training data very well, but misses the general rule behind the data. In the PAC model, overfitting is explained by a too rich hypothesis class. If the learner can choose a model that fits perfectly the training data - it will do so, ignoring the fact that the chosen model will possibly not be able to explain new data. Say for example that the hypothesis class consists of all the functions from the sample space to the labels space. A na\"{i}ve learner might choose a classifier that handles all the training points well, whereas any unknown sample is classified as $+1$. This selection might obviously have erroneous results. In the spirit of this idea, the PAC theory bounds the difference between the empirical and the actual risk using a combinatorial measure of the hypothesis class complexity, named \emph{VC dimension}. For a detailed review see \cite{vapnik1999}.

A common solution for this problem is to add a regularization term to the objective of the minimization problem. Usually, a norm of the classifier is taken as a regularization term. From the statistical learning theory's point of view, the regularization restricts the complexity of the model, and by that controls the difference between the training and testing error (\cite{Smola98,Evg00,Bartlett02}). The idea of minimizing the complexity of the model is not unique to the statistical theory, and may be traced back to the Ocaam's razor principle: the simplest hypothesis that explains the phenomenon is likely to be the correct one. Another way to understand the regularization term, is as a means to introduce prior knowledge.

\section {Support Vector Machines}

In support vector machine (SVM), the loss measure at hand is the hinge-loss
\be
\ell_{hinge}(\xx^m,y^m;\ww)=[1-y^m\ww^T\xx^m]_+ \nonumber
\ee
$\ell_{hinge}$ is a surrogate loss function, in the sense that it upper-bounds the zero-one loss. Furthermore, $\ell_{hinge}$ is convex, which makes it a far more convenient objective for numerical optimization than the zero-one loss. Note that the hinge loss intorduces penalty when the classifier correctly predicts the label of a sample, but does so with too little margin, i.e. $\ww^T\xx^m<1$. The penalty on a wrong classification is linear in the distance of the sample from the hyperplane.

\begin{figure}[h!]
  \centering
    \includegraphics[scale=0.4]{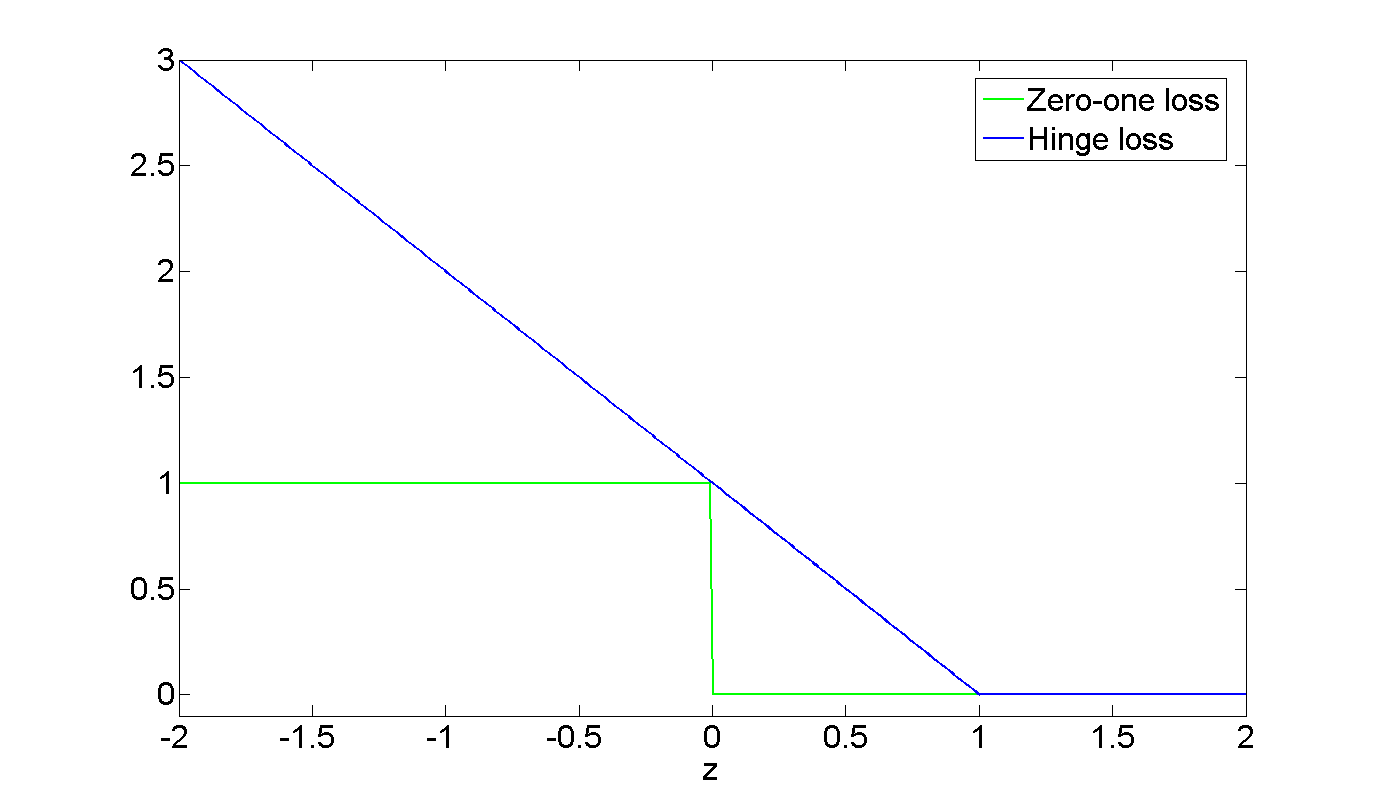}
  \caption{The hinge loss is a convex surrogate to the zero-one loss. }
	\label{fig:svm}
\end{figure}

As discussed, optimizing the sum of the losses solely may result in poor generalization performace. The SVM solution is to add an $L_2$ regularization term. The geometrical intuition behind this term is the following: The distance between the point $\xx^m$ and the hyperplane $\ww^T\xx=b$ is given by
\be
{{|\ww^T\xx^m-b|}\over{\|\ww\|}}
\ee
One may scale $\ww$ and $b$ in such a way that the point with the smallest margin (that is, the one closest to the hyperplane) will have $\|\ww^T\xx^m-b\|=1$. In that case, the bilateral margin is $\frac{2}{\|\ww\|}$ (see \figref{fig:svm}). This geometrical intuition, along with the fact that the hinge loss punishes too little margin, motivates the name \emph{Maximum Margin Classification} that was granted to SVM.
Hence, the SVM optimization task is
\be
\label{eq:introSVM}
\min_{\ww,b}{\frac{\lambda}{2}\|\ww\|^2+\sum_{m=1}^M{[1-y^m(\ww^T\xx^m-b)]_+}}
\ee

\begin{figure}[h!]
  \centering
    \includegraphics[scale=0.5]{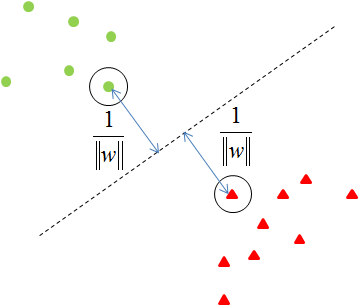}
  \caption{The bilateral margin is $\frac{2}{\|\ww\|}$. Thus, minimizing $\|\ww\|$ results in maximizing the margin.}
	\label{fig:svm}
\end{figure}
The parameter $\lambda$ controls the tradeoff between the training error and the margin of the classifier (cf. \secref{sec:brr}).

\section{Robustness}
The objective of the learning is to be able to classify new data. Thus, being robust to perturbations of the data is usually a desirable property for a classifier. In some cases, the training data and the testing data are sampled from different processes, which are similar to some extent but are not identical (\cite{Bi04}). This situation can happen also due to application specific issues, when new samples are sampled with reduced accuracy (for example, the training data may be collected with an expensive sensor, whereas cheaper sensors are deployed for actual use).

Even harder scenario is the one of learning in the presence of an adversary that may corrupt the training data, the testing data or both. The key step in order to formulate the robust learning task, is to model the action of the adversary, i.e., to define what is the family of perturbations that he may apply on the data points. In the Robust-SVM model, the adversary may apply a bounded additive distrubance, by displacing a sample point within a ball around it (\cite{SOCP06}). This case is referred to as box-uncertainty. \cite{Globerson06} assumed a different type of adversary. In their model, named FDROP, the adversary is allowed to delete a bounded number of features. This model results in more balanced classifiers, which are less likely to base their prediction only on a small subset of informative features.

Two issues usually repeat in robust learning formulations. The first one is the problem of the \emph{adversarial choice}. Most of the times, the first step in the analysis of the model is characterizing the exact action of the adversary on a specific data sample, given specific model parameters. The Robust-SVM adversary will choose to displace the point perpendicularly to the separating hyperplane. FDROP's adversary will delete the most informative features, i.e. those that have the maximal contribution to the dot product between the weights vector and the data point. The second issue is the restriction on the adversary's action. Regardless of the actual type of perturbation that the adversary uses, one needs to bound the extent to which it is applied. If no constraint is specified, the adversary will choose his action in such a way that the signal to noise ratio (SNR) will vanish, and the data will no longer carry any information. In the Robust-SVM formulation, the adversary is constrained to perform a displacement within a bounded ball. In the case of FDROP, no more than a pre-defined number of features may be deleted.

Note that robust formulations are closely related to the notion of \emph{consistency}. A classifier is said to be consistent, if close enough data points are predicted to have the same label. Different adversarial models befit different notions of distance. For example, the box-uncertainty model is related to the Euclidean metric and feature deletion suits the hamming distance.

It should be mentioned that robustness has quite a few meanings in the literature of statistics and machine learning. In this work, we use robustness in the sense of \emph{robust optimization} (RO), i.e. minimizing the worst-case loss under given circumstances.

\section {Between robustness and regularization}
\label{sec:brr}
The fact the robustness is related to regularization and generalization is not too surprising. Indeed, first equivalence results have been established for learning problems other than classification more than a decade ago (\cite{Gha97,XuCM08,Bishop94}). Recently, \cite{XuCaMa09} have proven the fact that the regularization employed by SVM is equivalent to a robust formulation. Specificly, they have shown that the following two formulations are equivalent
\bea
\min_{\ww,b}{\lambda\|\ww\|+\sum_{m=1}^M{[1-y^m(\ww^T\xx^m-b)]_+}} \nonumber \\
\min_{\ww,b}{
	\max_{\sum_m\|\delta_m\|^*\leq\lambda}
		{\sum_{m=1}^M{[1-y^m(\ww^T(\xx^m-\delta_m)-b)]_+}}
	}
\eea
where $\|\cdot\|^*$ is the dual-norm. This equivalence has a strong geometric interpertation, and sheds a new light on the function of the tuning parameter of SVM. Using the notion of robustness, a consistency result for SVM was given, without the use of VC or stability arguments. The novelty of that work stems from the fact that most previous works on robust classification were not aimed at relating robustness to regularization. Rather, the models were based on an already regularized SVM formulation, in which the loss measure was effectively modified.

\section {Our contribution}
In this work we adopt the idea of using robustness as a means to achieve generalization. We present a new robust-learning setup in which each data point is altered by a stochastic cloud centered on it. The loss is then evaluated as the expectation of an underlying loss on the cloud. The parameters of this cloud's distribution are chosen in an adversarial fashion. We analyze the case in which the adversary is restricted to choose a Gaussin cloud with a trace-bounded covariance matrix. Then we show that this formulation culminates in a smooth upper-approximation of the hinge loss, which gets tighter as the cloud around each data sample shrinks. This loss function can be shown to have a convex perspective structure. By deriving the dual problem, we are able to demonstrate a method of generating new smooth loss functions.
Our algorithmic approach is to directly solve the primal problem. We show that this yields a learning algorithm which generalizes as well as SVM on synthetic as well as real data. Generalizations to the non-linear and multiclass cases are given.

\section{Related work}
Other works have incorporated a noise model into the learning setup. For example, \cite{JBP08} have warped the data points with ellipsoids. \cite{IDIAP} have shown how to train classifiers for distributions. Similar to what we do in this work, they use tails of distributions in their derivation. Their work, however, treated each data class as a distribution, whereas in this work we attach a noise distribution for each data point separately. \cite{SOCP04,SOCP06} have employed second order cone programming (SOCP) methods in order to handle the uncertainty in the data. \cite{Molecular04} have assumed stochastic clouds instead of discrete points, as we do, but they did not try to minimize the expectation of the loss function over the cloud. Instead, their idea was to incorporate the idea with the soft margin framework. \cite{Bi04} have tried to learn a better classifier by presenting the learning algorithm 'more reasonable' samples. We elaborate on this model in \appref{app:singAlgs}.

Smooth loss function were studied by \cite{Zhang,Chap07}. Analysis of methods for Solving SVM and SVM-like problems using the primal formulation was done by \cite{Pegasos,Chap07}.

\mbox{}\newline
The rest of this document is organized as follows: in \chapref{section-GURU} we present our framework formally, derive the explicit form of the smooth loss function and devise an algorithm that finds the optimal classifier. In \chapref{section-dual} we derive a dual formulation for the problem, and point out that our model may be generalized for other loss functions. In \chapref{section-kernels} we apply the kernel trick and devise a method for training non-linear classifiers in the same cost as for the linear kernel. \chapref{section-multi} contatins a generalization of the binary algorithm for the multiclass case. At last, in \chapref{section-discussion} we discuss the contributions presented in this work and mention possible directions for future work.
In \appref{app:singAlgs} we discuss a far more basic version of resistance to noise. The results of the first section therein are not original and presented here only for the sake of logical order. The next section contains a simple generalization for the multiclass case. \appref{app:diagCov} gives the solution to the adversarial choice problem for an adversary that is restricted to spread the noise along the primary axes. At last, in \appref{app:mhinge} we explain why we find the usual multiclass hinge loss inapplicable in our framework.

\chapter{Gaussian Robust Classification}
\label{section-GURU}
In this work we take the approach of robust optimization (RO). Accordingly, we present a min-max learning framework, in which the learner strives to minimize the loss, whereas the adversary tries to maximize it. The model that we introduce in this chapter has two layers of 'robustness'. Firstly, we use the min-max robustness, which lays in the foundations of RO. Secondly, we effectively enhance the training dataset by taking into consideration all the possible outputs of the adversarial perturbation. 
More concretely, we alter each training sample with a stochastic cloud. The shape of this cloud is chosen by the adversary from a pre-determined family of distributions. The spreading of the samples should be understood as adding noise, where different disturbances take place with different probability. The loss on each sample is finally computed as the expectation of an underlying loss on the respective cloud.

\section{Problem Formulation}
In this section we formally describe the model we investigate in the work. We take the hinge-loss as the underlying loss function, and build the learning framework on top of it. We then show that the new framewok we introduce is equivalent to an unconstrained minimization of an effective loss function.

Recall that the hinge loss is defined
\be
\ell_{hinge}(\xx^m,y^m;\ww) = [1-y^m\ww^T\xx^m]_+
\ee
We introduce the expected hinge loss
\be
\ell_{hinge}^{\mathbb{E}}(\xx^m,y^m;\ww,\mathcal{D}) = \mathbb{E}_{\nn\sim\mathcal{D}}\left[1-y^m\ww^T(\xx^m+\nn)\right]_+
\ee
where $\mathcal{D}$ is a predefined noise distribution over the sample space. The optimization problem for learning a classifier w.r.t. the expected hinge loss is thus
\be
\min_{\ww}{\sum_{m=1}^M{\ell_{hinge}^{\mathbb{E}}(\xx^m,y^m;\ww,\mathcal{D})}}
\ee
Granting an adversary the ability to choose the noise distribution, we end up with the following formulation
\be
\min_{\ww}{\max_{\mathcal{D}_1\times\mathcal{D}_2\times...\times\mathcal{D}_M\in
\mathcal{C}_1\times\mathcal{C}_2\times...\times\mathcal{C}_M}
	{\sum_{m=1}^M{\ell_{hinge}^{\mathbb{E}}(\xx^m,y^m;\ww,\mathcal{D}_m)}}}
\ee
where $\mathcal{C}_m$ is the set of allowed noise distributions for the $m^{\text{th}}$ sample. In order for the adversarial optimization to be meaningful, all $\mathcal{C}_m$'s should have a 'bounded' nature. We now alter the order of maximization and summation, and write
\be
\label{eq:generalTask}
\min_{\ww}{{\sum_{m=1}^M{\max_{\mathcal{D}_m\in\mathcal{C}_m}{\ell_{hinge}^{\mathbb{E}}(\xx^m,y^m;\ww,\mathcal{D}_m)}}}}
\ee
At last, we observe that the optimization task at hand is nothing else than optimizing the effective loss function
\be
\label{eq:robEhinge}
\ell_{hinge}^{rob}(\xx^m,y^m;\ww,\mathcal{D}) = \max_{\mathcal{D}\in\mathcal{C}}\mathbb{E}_{\nn\sim\mathcal{D}}\left[1-y^m\ww^T(\xx^m+\nn)\right]_+
\ee
We refer to $\ell_{hinge}^{rob}(\xx^m,y^m;\ww,\mathcal{D})$ as the expected robust hinge loss.
\section{The adversarial Choice}
\eqref{eq:generalTask} presents the general noise-robust formulation. In the following, we will derive an explicit loss function for a specific collection of noise distributions. We focus on the case in which the adversary is constrained to spread a Gaussian noise, having a trace bounded covariance-matrix. The motivation behind this constraint is physical. When a noise is modeled with a distribution, its covariance is considered as its power. Thus, by constraining the sum of the eigenvalues of the covariance matrix we bound the power that the adversary can spread. The Gaussian noise is the worst case noise, in the sense that amongst all distributions with a certain poer bound it has the maximal entropy.

Using the notations of the previous section, we specify the restriction on the adversary as
$$\mathcal{C} =\mathcal{C}_\beta = \left\{\mathcal{D}\sim\mathcal{N}(\bf{0},\Sigma)|\Sigma\in\Lambda_\beta\right\}$$ 
where $\Lambda_\beta=\left\{\Sigma\in\text{PSD}|Tr(\Sigma)\leq\beta\right\}$,
i.e. Gaussian distributions having the zero vector as mean and a covariance matrix with a bounded sum of eigenvalues.

In the next couple of sections we will characterize the adversarial choice of the covariance matrix and derive an explicit loss function.

\subsection{The structure of the loss function}
\label{guru-loss-structure}
The following paragraphs are rather technical. For later use, we explicitly perform the integration of the robust hinge loss function. We then prove a monotony property of the integrated loss. This property will help us analyze the nature of the adversarial choice in our case. The key observation throughout the derivation is that the multivarite expectation can be transformed to a univariate problem.

We plug the notations that were introduced above into \eqref{eq:robEhinge} and get:
\be
\ell_{hinge}^{rob}(\xx^m,y^m;\ww,\Sigma) = \max_{\Sigma\in \Lambda_\beta} c|\Sigma|^{-{1\over 2}}\int e^{-{1\over 2}\nn^T\Sigma^{-1}\nn}\left[ 1-y^m \ww^T(\xx^m+\nn) \right]_{+} d\nn  
\ee
where $c=(2\pi)^{-d/2}$ is the normalization constant. This is equivalent to:
\be
\label{eq:rehX}
\ell_{hinge}^{rob}(\xx^m,y^m;\ww,\Sigma) =  \max_{\Sigma\in \Lambda_\beta} c|\Sigma|^{-{1\over 2}}\int e^{-{1\over 2}\nn^T\Sigma^{-1}\nn}\left[ 1-y^m \ww^T\xx^m-y^m\ww^T\nn \right]_{+} d\nn  
\ee

As a first step in the analysis of the expected robust hinge loss, we shall handle the quantity
\be
\label{eq:rehIntegrand}
Q\stackrel{\mathrm{def}}{=}c|\Sigma|^{-{1\over 2}}\int e^{-{1\over 2}\nn^T\Sigma^{-1}\nn}\left[ 1-y^m \ww^T\xx^m-y^m\ww^T\nn \right]_{+} \nn
\ee

Note that the above only depends on $\nn$ via products of the form $\ww^T\nn$. Therefore, we define a new scalar variable $u = y^m\ww^T\nn$. \eqref{eq:rehIntegrand} can now be viewed as the expected value of $g(u) = [1-y^m\ww^T\xx^m - u]_+$. The moments of $u$ are
\bea
\mathbb{E}u &=& \mathbb{E}y^m\ww^T\nn = \nonumber \\
&=& y^m\ww^T\mathbb{E}\nn = 0
\eea
and
\bea
Var[u] &=& Var[y^m\ww^T\nn] = \nonumber\\
&=& y^m\ww^T Var[\nn]y^m\ww = \nonumber\\
&=& {\left(y^m\right)}^2\ww^T\Sigma\ww = \nonumber\\
&=& \ww^T\Sigma\ww
\eea
Thus we get
\be
\label{eq:massageIntegrand}
Q={1\over{\sqrt{2\pi\ww^T\Sigma\ww}}}\int e^{-{u^2\over 2\ww^T\Sigma\ww}}\left[ 1-y^m \ww^T\xx^m-u \right]_{+} du  
\ee
Define ${\erf}(t)={1\over\sqrt{2\pi}}\int_{-\infty}^t{\exp(-{1\over{2}}z^2)dz}$. In addition, denote $\sigma^2(\ww,\Sigma) = {\ww^T\Sigma\ww}$. The following proposition holds.
\pro
\label{pro:integration}
$Q=\left(1-y^m\ww^T\xx^m\right) \erf\left({{1-y^m\ww^T\xx^m}\over{\sqrt{\sigma^2(\ww,\Sigma)}}}\right)
+{\sqrt{\sigma^2(\ww,\Sigma)\over 2\pi}}\exp{\left(-{\left(1-y^m\ww^T\xx^m\right)^2\over {2\sigma^2(\ww,\Sigma)}}\right)}$
\epro

\prf
We conduct a direct computation:
\bea
Q &=& {1\over{\sqrt{2\pi\sigma^2(\ww,\Sigma)}}}\int e^{-{u^2\over 2\sigma^2(\ww,\Sigma)}}\left[ 1-y^m \ww^T\xx^m-u \right]_{+} du = \nonumber\\
&=& {1\over{\sqrt{2\pi\sigma^2(\ww,\Sigma)}}}\int_{-\infty}^{1-y^m \ww^T\xx^m} e^{-{u^2\over 2\sigma^2(\ww,\Sigma)}}\left( 1-y^m \ww^T\xx^m-u \right) du = \nonumber\\
&=& {1\over{\sqrt{2\pi\sigma^2(\ww,\Sigma)}}}\left(\left( 1-y^m \ww^T\xx^m \right)\int_{-\infty}^{1-y^m \ww^T\xx^m} e^{-{u^2\over 2\sigma^2(\ww,\Sigma)}} du \right.\nonumber\\
&\qquad&\qquad - \left.\int_{-\infty}^{1-y^m \ww^T\xx^m} ue^{-{u^2\over 2\sigma^2(\ww,\Sigma)}} du\right) = \nonumber\\
&=& {\left( 1-y^m \ww^T\xx^m \right)}\erf{\left({{ 1-y^m \ww^T\xx^m} \over{\sqrt{\sigma^2(\ww,\Sigma)}}}\right)} \nonumber\\
&\qquad&\qquad-{1\over{\sqrt{2\pi\sigma^2(\ww,\Sigma)}}}\int_{-\infty}^{1-y^m \ww^T\xx^m} ue^{-{u^2\over 2\sigma^2(\ww,\Sigma)}} du \nonumber
\eea
By using the variable substitution theorem and observing that the remaining integrand is an odd function (thus the identity $\int_{-\infty}^{t}{\text{odd}}=\int_{-\infty}^{-t}{\text{odd}}$ holds), we conclude that
\be
Q=\left(1-y^m\ww^T\xx^m\right) \erf\left({{1-y^m\ww^T\xx^m}\over{\sqrt{\sigma^2(\ww,\Sigma)}}}\right)
+{\sqrt{\sigma^2(\ww,\Sigma)\over 2\pi}}\exp{\left(-{\left(1-y^m\ww^T\xx^m\right)^2\over {2\sigma^2(\ww,\Sigma)}}\right)}
\ee
\eprf
\newline
Let us establish the following simple property of $Q$.
\lem
\label{lem:mono}
$Q$ is monotone-increasing in $\sigma^2$.
\elem
\prf
The fundamental theorem of calculus yields that
\be
{d\over{dt}}\erf(t) = {1\over{\stp}}\exp{\left(-{t^2\over{2}}\right)}
\ee
Using the chain rule we compute
\be
{dQ\over{d\sigma^2}} = {1\over{2\stp\sqrt{\sigma^2}}}\exp{\left(-{\left(1-y^m\ww^T\xx^m\right)^2\over {2\sigma^2(\ww,\Sigma)}}\right)}
\ee
It is evident that for all $\sigma^2\geq 0$ \be{dQ\over{d\sigma^2}}\geq 0\ee
i.e. $Q$ is monotone-increasing in $\sigma^2$.

\eprf
\subsection{The optimal covariance matrix subject to a trace constraint}	

We will now focus on finding the optimal adversary, i.e., performing the maximization of \eqref{eq:rehX} over the range of allowed covariance matrices.
The next theorem specifies which covariance matrix attains the worst-case loss. In out terminology, refer to this result as the \emph{adversarial choice}.

\begin{figure}[h!]
  \centering
    \includegraphics[scale=0.5]{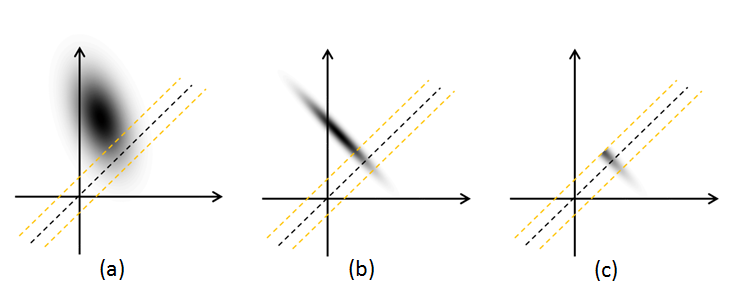}
  \caption{(a) Replacing the sample point with a Gaussian cloud. (b) The optimal noise direction is orthogonal to the separating hyperplane. (c) The expected robust hinge loss only considers the tail of the distribution, i.e. the points that suffer a margin error.}
	\label{fig:BinAdverChoice}
\end{figure}

\thm
\label{thm:bacThm}
The optimal $\Sigma$ in \eqref{eq:rehX} is given by $\Sigma^* = \beta{{\ww\ww^T}\over{\|\ww\|^2}}$ where the optimization is done over $\Sigma\in \Lambda_\beta$.
\ethm
Before actually proving the theorem, we will give some geometric intuition. The idea behind the expected loss is to replace the original sample point with a Gaussian cloud centered at the original point (\figref{fig:BinAdverChoice}a). Consider an arbitrary displacement $\hat{\xx^m} = \xx^m + \nn$. For fixed $\ww$, $\nn$ can be written as $\nn = \nn_{\|}+\nn_\bot$. The relevant quantity is $\ww^T\hat{\xx^m} = \ww^T\xx^m+\ww^T\nn_{\|}$, that is, the orthogonal component does not have any effect. Accordingly, it makes sense that the optimal noise direction is orthogonal to the separating hyperplane, i.e. parallel to the vector $\ww$ (see \figref{fig:BinAdverChoice}b).

The proof of \thmref{thm:bacThm} applies simple algebraic results to establish this result rigorously. \newline

\prf
Plugging \proref{pro:integration} into \eqref{eq:rehX} we get
\bea
\ell_{hinge}^{rob}(\xx^m,y^m;\ww,\Sigma) &=& \max_{\Sigma\in\Lambda_\beta}\left[\left(1-y^m\ww^T\xx^m\right) \erf\left({{1-y^m\ww^T\xx^m}\over{\sqrt{\sigma^2(\ww,\Sigma)}}}\right)\right.\nonumber\\
&\qquad&\qquad+\left.{\sqrt{\sigma^2(\ww,\Sigma)\over 2\pi}}\exp{\left(-{\left(1-y^m\ww^T\xx^m\right)^2\over {2\sigma^2(\ww,\Sigma)}}\right)}\right] \eea

The above depends on $\Sigma$ only via $\sigma^2(\ww,\Sigma)$. According to \lemref{lem:mono}, the objective is monotone increasing in $\sigma^2$. Therefore, the adversary would like to choose $\Sigma$ so that $\sigma^2(\ww,\Sigma)$ is maximized. By applying the Cauchy-Schwartz inequality, we conclude that the maximum value of $\sigma^2(\ww,\Sigma)$ is $\lambda_{\max}(\Sigma)\|\ww\|^2$.
For all $\Sigma\in\Lambda\beta$ it holds that $Tr(\Sigma)\leq \beta$. Since all of the eigenvalues are positive, it holds that $\lambda_{\max}\leq\beta$ as well. Consider the candidate solution $\Sigma_0 = \beta{{\ww\ww^T}\over{\|\ww\|^2}}$. Since $\sigma^2(\ww,\Sigma_0) = \beta\|\ww\|^2$, this selection attains the maximum. Note that this covariance matrix reflects the fact that the adversarial choice would be to spread the noise parallel to the separator.
\eprf

\section{A smooth loss function}
In the previous sections we have done the technical computations needed in order to derive the robust hinge loss explicitly, and found the optimal covariance matrix. In the following we will put these results together, and present an explicit formulation of the loss function resulting from our model. In addition, it is shown that our robust loss can be represented as a perspective of a scalar smooth approximation of the hinge loss.  By analyzing this function we are able to gain a better understanding of $\lrh$. We conclude this section by showing that our loss function is a smooth convex upper-approximation of the hinge-loss. When the 'diameter' of the noise cloud is shrunk, $\lrh$ coincides with the hinge-loss.

We devote a notation for the result of \proref{pro:integration}
\bea
\label{eq:lossAfterInt}
L(\xx^m,y^m;\ww,\sigma^2) &=& \left(1-y^m\ww^T\xx^m\right) \erf\left({{1-y^m\ww^T\xx^m}\over{\sigma}}\right)\nonumber\\
&+& {\sigma\over\sqrt{2\pi}}\exp{\left(-{\left(1-y^m\ww^T\xx^m\right)^2\over {2\sigma^2}}\right)}
\eea

By combining the above equation with the result of \thmref{thm:bacThm} we conclude
\bea
\label{eq:lossResult}
\ell_{hinge}^{rob}(\xx^m,y^m;\ww,\beta) &=&\left(1-y^m\ww^T\xx^m\right) \erf\left({{1-y^m\ww^T\xx^m}\over{\sqrt{\beta}\|\ww\|}}\right)\nonumber\\
&+&{\sqrt{\beta}\|\ww\|\over\sqrt{2\pi}}\exp{\left(-{\left(1-y^m\ww^T\xx^m\right)^2\over {2\beta\|\ww\|^2}}\right)}
\eea
$\beta$ has the meaning of statistical variance, and therefore in the following we will replace it with $\sigma^2$ (not to be confused with $\sigma^2(\ww,\Sigma)$).
In order to understand the nature of the loss function we have defined, it is suggestive to define
\be
f(z) = z\erf(z)+{1\over{\sqrt{2\pi}}}e^{-{z^2\over 2}}
\ee
Using $f$, the robust expected hinge loss can be written as
\be
\ell_{hinge}^{rob}(\xx^m,y^m;\ww,\ssq) = \sigma\|\ww\| f\left(\frac{1-y^m\ww^T\xx^m}{\sigma\|\ww\|}\right)
\ee
A direct computation shows that
\be
\label{eq:fDeriv}
{df\over{dz}} = \erf(z) + {z\over{\sqrt{2\pi}}}e^{-{z^2\over 2}} - {z \over{\sqrt{2\pi}}}e^{-{z^2\over 2}} = \erf(z)
\ee
We are now ready to prove a simple yet fundamental property of $f$.
\thm
\label{thm:fProp}
$f$ is a smooth strictly-convex upper-approximation of the hinge loss.
\ethm

\begin{figure}[h!]
  \centering
    \includegraphics[scale=0.4]{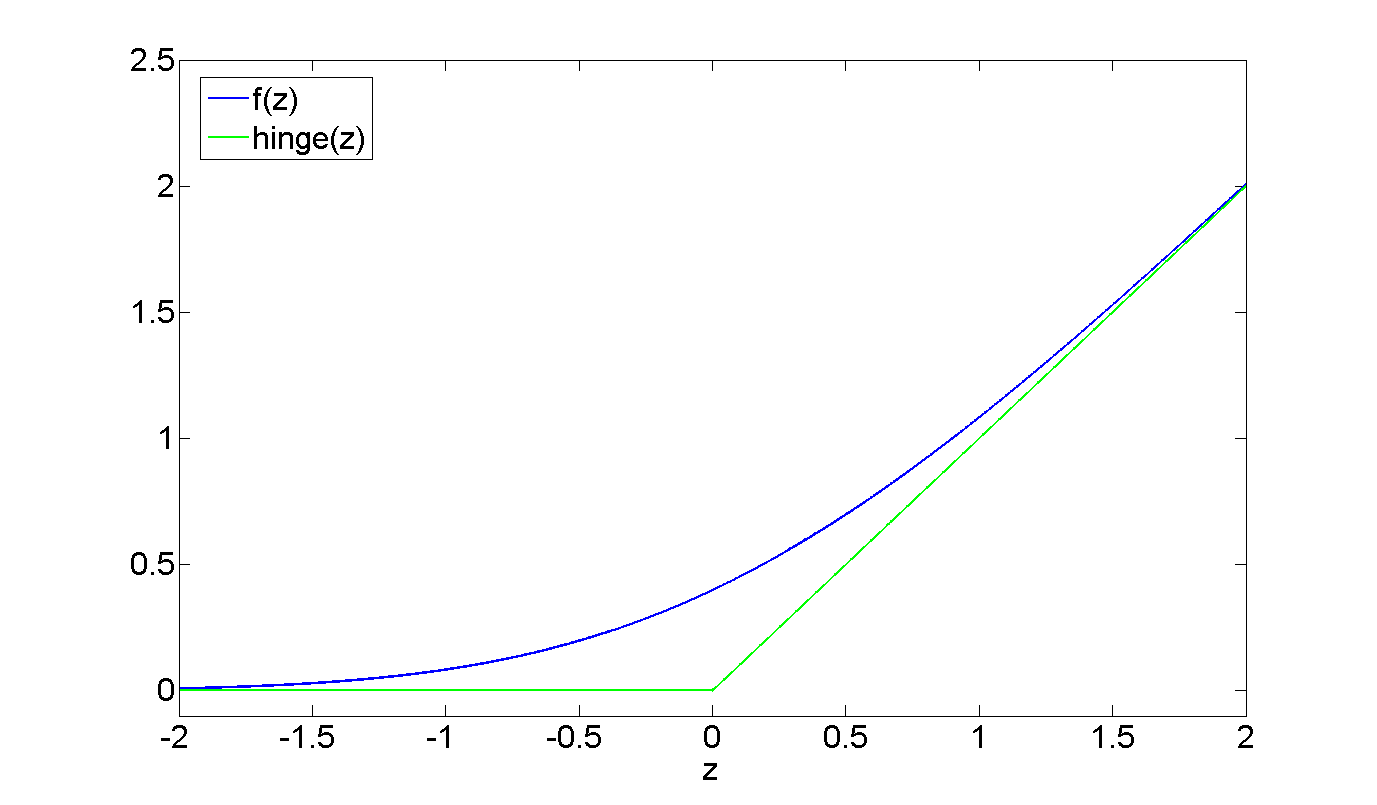}
  \caption{The function $f$ is a smooth approximation of the hinge loss}
\end{figure}

\prf
Denote the hinge loss $h(z) = [z]_+$.
We must show that
\begin{enumerate}
\item $f$ is strictly-convex
\item $f(z)\geq h(z)$
\item $\lim_{z\rightarrow -\infty}f(z)-h(z) = 0$
\item $\lim_{z\rightarrow \infty}f(z)-h(z) = 0$
\end{enumerate}
Differentiating \eqref{eq:fDeriv} once again, we get
\be
{d^2 f\over{dz^2}} = {1\over{\stp}}e^{-{z^2\over 2}}
\ee
which is clearly positive for all $z$. Thus, $f$ is strictly-convex. \newline
For the upper bound property, notice that $f(z)\geq 0$ for all $z\in\mathbb{R}$. Hence, for $z<0$ we simply have $f(z)>0=h(z)$. For the complementary case, denote the difference function over the positives $\delta(z) = f(z)-h(z) = z\erf(z)+{1\over{\sqrt{2\pi}}}e^{-{z^2\over 2}}-z$. Using \eqref{eq:fDeriv} we obtain
\be
{dh\over{dz}} = \erf(z)-1
\ee
It can be easily seen that ${dh\over{dz}}<0$, i.e. $h$ is monotone decreasing. Observe that $\delta(0) = {1\over{\stp}}$ and $\lim_{z\rightarrow\infty}\delta(z)=0$. Since all the functions involved are continous, we conclude that for $z\geq 0$ it holds that $h(z)\leq f(z) \leq h(z) + {1\over{\stp}}$. Altogether, we have established the upper bound property. \newline
For the asymptote at $z\rightarrow -\infty$, observe that from l'Hopital \be\lim_{z\rightarrow -\infty}z\erf(z)=\lim_{z\rightarrow -\infty}\stp e^{-{z^2\over 2}} = 0\nonumber\ee
Since the exponent in the right summand of $f$ decays as well, we have that at $z\rightarrow -\infty$ both $f(z)$ and $h(z)$ coincide on $z=0$. \newline
For the asymptote at $z\rightarrow \infty$, we must show that $f$ asymptotically coincide with the linear function $z$. To this end, let us write $f(z)-z = z\left(\erf(z)-1\right)+{1\over{\stp}}e^{-{z^2\over 2}}$. Applying l'Hopital rule along with the asymptotic behavior of the exponent, we deduce that $\lim_{z\rightarrow \infty}f(z)-z = 0$, as desired.
\eprf \newline

Next, we will analyze the relation between $f$ and $\lrh$.
\dfn  \textbf{Perspective of a function} (from \cite{BoydOpt} $3.2.6$). \newline
If $f:\mathbb{R}^n\rightarrow\mathbb{R}$, then the \emph{perspective} of $f$ is the function $g:\mathbb{R}^{n+1}\rightarrow\mathbb{R}$ defined by
\be
g(t,x)=tf\left({x\over{t}}\right) \nonumber
\ee
with domain
\be
\text{dom}(g) = \left\{(x,t)\Big |{x\over{t}}\in\text{dom}(f),t>0\right\} \nonumber
\ee
\edfn
The following lemma if useful. For a proof see \cite{BoydOpt} $3.2.6$, e.g.
\lem
\label{lem:persIsCvx}
If $f$ is convex (concave), then its perspective is convex (concave) as well.
\elem
Define the function
\be
g(a,b) = af\left({b\over{a}}\right)
\ee
\lemref{lem:persIsCvx} implies that $g\left(\sigma\|\ww\|,1-y^m\ww^T\xx^m\right)$ is jointly convex in both its arguments. In order to establish the strict-convexity of $\lrh$ in $\ww$, we need a more powerful tool. Consider the following lemma (\cite{BoydOpt} $3.2.4$)
\lem
\label{lem:compCvx}
Let $h:\mathbb{R}^k\rightarrow\mathbb{R}$, $g_i:\mathbb{R}^n\rightarrow\mathbb{R}$. Consider the function $f(x)=h\left(g(x)\right)=h\left(g_1(x),g_2(x),...,g_k(x)\right)$. Then, $f$ is convex if $h$ is convex, $h$ is nondecreasing in each argument, and $g_i$ are convex.
\elem
This lemma can be easily generalized to the case of strictly-convex functions. The proof is identical to that of the original version, thus will be skipped. \newline
We are now ready to prove the following theorem
\thm
\label{thm:actualCvx}
$\lrh$ is strictly-convex in $\ww$.
\ethm

\prf
From \lemref{lem:persIsCvx}, $g$ is convex. In additon, $g$ is nondecreasing in each of its arguments. To see that, observe that
\bea
{dg\over{da}}&=&f\left({b\over{a}}\right)-{b\over{a}}f'\left({b\over{a}}\right)= {1\over{\stp}}\exp\left(-{1\over{s}}{b^2\over{a^2}}\right)\nonumber\\
{dg\over{db}} &=& f'\left({b\over{a}}\right) = \erf\left({b\over{a}}\right)\nonumber
\eea
which are both strictly positive.\newline
$\sigma\|\ww\|$ and $\left(1-y^m\ww^T\xx^m\right)$ are both convex in $\ww$, thus we conclude by applying \lemref{lem:compCvx}.
\eprf \newline

The next theorem explores some of the other properties of the loss function we have defined:
\thm
$\lrh$ is an upper-approximation to the hinge loss. Furthermore, when $\sigma^2\rightarrow 0$, the loss function $\lrh$ coincides with the hinge loss.
\ethm
\prf
For the upper bound property, we apply \thmref{thm:fProp}:
\bea
\sigma\|\ww\|f\left({1-y^i\ww^T\xx^i}\over {\sigma\|\ww\|}\right)&\geq&
\sigma\|\ww\|h\left({1-y^i\ww^T\xx^i}\over {\sigma\|\ww\|}\right)\nonumber \\ &=& 
\sigma\|\ww\|\left[{(1-y^i\ww^T\xx^i}\over {\sigma\|\ww\|}\right]_+ \nonumber \\
&=&
\left[1-y^i\ww^T\xx^i\right]_+
\nonumber
\eea
For the second part of the theorem, let us first observe that
\be
{\sigma\over\sqrt{2\pi}}\exp{\left(-{\left(1-y^m\ww^T\xx^m\right)^2\over {2\sigma^2}}\right)}\rightarrow 0
\ee
as a multiplication of two vanishing factors at $\sigma\rightarrow 0$. We consider two cases:
\begin{enumerate}
\item \textbf{$1-y^m\ww^T\xx^m\geq 0$}. Observe that
\be
\erf\left({{1-y^m\ww^T\xx^m}\over{\sigma\|\ww\|}}\right)\rightarrow\erf(\infty) = 1 \nonumber
\ee
Thus, $\lrh(\xx^m,y^m;\ww,\sigma^2)\rightarrow 1-y^m\ww^T\xx^m$.
\item \textbf{$1-y^m\ww^T\xx^m < 0$}. In this case
\be
\erf\left({{1-y^m\ww^T\xx^m}\over{\sigma\|\ww\|}}\right)\rightarrow\erf(-\infty) = 0 \nonumber
\ee
Thus, $\lrh(\xx^m,y^m;\ww,\sigma^2)\rightarrow 0$
\end{enumerate}
Altogether, we have shown that when $\sigma\rightarrow 0$, $\lrh(\xx^m,y^m;\ww,\sigma^2)\rightarrow\left[1-y^m\ww^T\xx^m\right]_+$.
\eprf

Observe that at $\ww=\boldsymbol{0}$ the loss function is not continuous. The discontinuity is removable, however, so this issue does not pose any problem.

\begin{figure}[h!]
  \centering
    \includegraphics[scale=0.4]{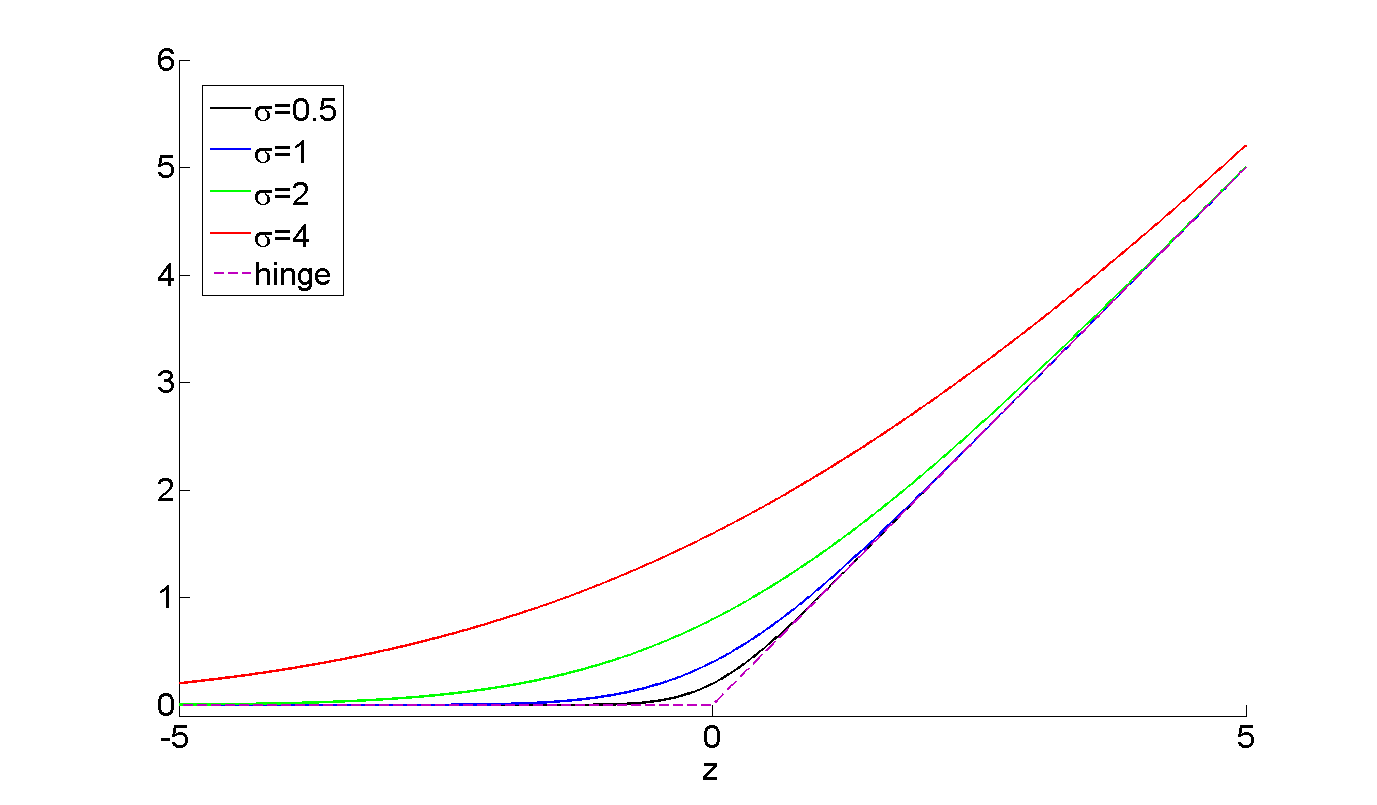}
  \caption{$\lrh$ is a convex upper-approximation to the hinge loss. As $\sigma$ tends to $0$, $\lrh$ tends to the hinge. In all of the graphs, the norm $\|\ww\|$ was set to 1.}
\end{figure}

The norm of the classifier $\|\ww\|$ always appears in a multiplication with $\sigma$. Thus, we observe that it has a similar function. Namely, it controls the tightness of the approximation of the smooth loss function to the hinge. Since $\sigma$ is pre-determined, the optimal norm should reflect some kind of compensation. We thus conjecture that there exist a inverse ratio between $\sigma$ and the optimal norm (cf. \chapref{section-kernels}).

At last, it should be noted that this loss smooth function can be viewed as a multiplicative regularized loss.



\section{GURU: a primal algorithm}
\label{sec:guru_alg}
We are now ready to devise an algorithm that solves our learning problem. In this section we describe a stochastic gradient descent (SGD) method that minimizes the strictly-convex loss function at hand. A convergence result for the algorithm stems from general properties of SGD that were studied extensively (see \cite{pegasosFull,onlineLWK,Zhang,Nedic,SGDformal}, e.g.).

Plugging the robust hinge loss function we have derived (\eqref{eq:lossResult}) into the original optimization task (\eqref{eq:generalTask}), we get
\be
\label{eq:forGURU}
\min_{\ww}{\sum_{m=1}^M{
	\left(1-y^m\ww^T\xx^m\right)
		\erf\left({{1-y^m\ww^T\xx^m}\over{\sigma\|\ww\|}}\right)
	+{\sigma\|\ww\|\over{\sqrt{2\pi}}}
		\exp{\left(-{\left(1-y^m\ww^T\xx^m\right)^2
			\over{2\sigma^2\|\ww\|^2}}\right)
		}
	}
}
\ee
This formulation is a convex unconstrained minimization task. One very natural approach for solving this kind of task is using the family gradient descent methods. Denote the objective of the optimization as
\be
G(\ww)=\sum{g_i(\ww)}
\ee
In batch gradient descent, in each step the algorithm updates
\be
\ww\leftarrow\ww-\eta\nabla G(\ww)
\ee
In stochastic gradient methods the gradient is approximated as the gradient of one of the summands. Thus, the algorithm first randomizes an index $i$, then updates
\be
\ww\leftarrow\ww-\eta\nabla g_i(\ww)
\ee
where $\eta$ is the learning rate. The stochastic version suits settings of online learning, in which the learner is presented one training sample at a time. It has been suggested that using the stochastic version yields better generalization performance in learning tasks (\cite{Amari98,Bot03}).

Our algorithm, named GURU (\textbf{G}a\textbf{U}ssian \textbf{R}ob\textbf{U}st), optimizes \eqref{eq:forGURU} using an SGD procedure. (For a full treatment see, e.g. Boyd (ref).)\newline
In order to derive the update formula, one should first calculate the gradient of the loss function. A straight forward computation yields
\be
\label{eq:lossDeriv}
\nabla_{\ww} {\lrh(\xx^i,y^i;\ww,\ssq)}=
-y^i\xx^i \erf\left({1-y^i\ww^T\xx^i}\over {\sigma\|\ww\|}\right)+{\sigma\ww\over\sqrt{2\pi}\|\ww\|}\exp\left(-{(1-y^i\ww^T\xx^i)^2\over 2\sigma^2\|\ww\|^2}\right)
\ee

We therefore suggest the following SGD procedure

\begin{algorithm}[H]
\SetLine
\KwData{Training set $\cal S$, learning rate $\eta_0$, accuracy $\epsilon$}
\KwResult{$\ww$}
\caption{GURU($\cal S$,$\eta_0$,$\epsilon$)}
$\ww \leftarrow \boldsymbol{0}$\;
\While{$\Delta L \geq\epsilon$} {
	$m \leftarrow rand(M)$\;
	$\ww \leftarrow \ww-{\eta_0\over \sqrt{t}}\nabla_{\ww}\lrh(\xx^m,y^m;w,\ssq)$\;
}
\Return $\ww$\;
\end{algorithm}

For convergence results see \cite{Nedic}. For a full treatment, see \cite{breBook}, chapter $8$.

\section {Experiments}
\label{sec:bExp}
\begin{figure}[tbp]
  \centering
    \includegraphics[scale=0.6]{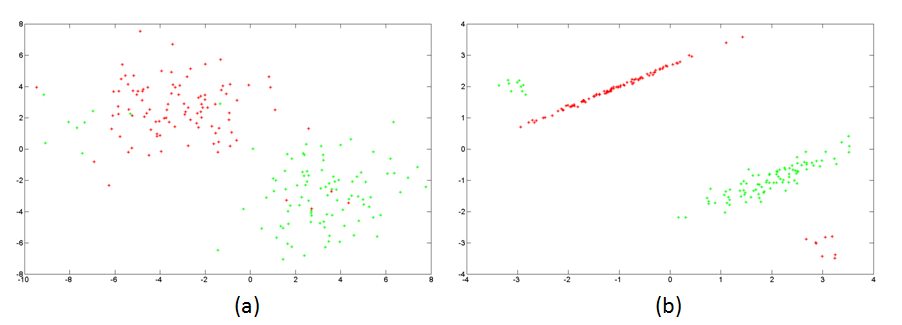}
  \caption{(a) Gaussian data. (b) Narrow Gaussian with outliers.}
	\label{fig:linData}
\end{figure}

In this section we present experimental results that demonstrate the fact that GURU generalizes as well as SVM. Experiments were carried out on two toy problems (see \figref{fig:linData} for a visualization), USPS handwritten digits classification (3 vs. 5, 5 vs. 8 and 7 vs. 9) and a couple of UCI databases (\cite{UCI}). The sizes of the data sets are detailed in \tabref{tab:binSizes}.

\begin{table}
	\begin{center}
	    \begin{tabular}{ p{1.2cm} p{1.5cm} p{1.5cm} p{1.5cm} p{1.5cm} }
	    \hline
	    Name & \#Training samples & \#Cross-validation samples & \#Test samples & \#features \\ \hline
	    Toy(a) & 200 & 200 & 200 & 2\\
	    Toy(b) & 200 & 200 & 200 & 2\\
		 Ionosp-here & 100 & 100 & 152 & 34\\
		 diabetes & 200 & 100 & 468 & 8\\
		 splice 1 vs. 2 & 500  & 400 & 635 & 60\\
	    USPS 3 vs. 5 & 800 & 700 & 700 & 256\\
	    USPS 5 vs. 8 & 800 & 700 & 700 & 256\\
	    USPS 7 vs. 9 & 800 & 700 & 700 & 256\\ \hline
	    \end{tabular}
		\caption{Description of the databased used in the binary case}
		\label{tab:binSizes}
	\end{center}
\end{table}

We have trianed GURU for $\sigma$ values varying from $2^{-20}$ to $2^{20}$, with exponential jumps. The learning rate was tuned empirically (values between $4^{-10}$ to $4^{10}$ were tested). SVM was trained and tested using the SVM-light package. $\lambda$ values between $4^{-15}$ and $4^{15}$ were used. Note that in the SVM-light formulation, $\lambda$ multiples the loss and not the regularization term. Thus, the qualitative relation between $\lambda$ and $\sigma$ is roughly $\sigma\sim\frac{1}{\lambda}$. Parameter selection was done based on the cross-validation set, and performance was evaluated for the optimal parameter on a testing set. The results are summarized in \tabref{tab:binResults}.

\begin{table}
	\begin{center}
	    \begin{tabular}{ p{1.2cm} p{1.7cm} p{1.7cm} }
	    \hline
	    Name & GURU(\%) & SVM(\%) \\ \hline
	    Toy(a) & 92.5 & 92.5 \\
	    Toy(b) & 92 & 92 \\
		 Ionosp-here & 82.24 & 79.61 \\
		 diabetes & 67.52 & 67.31 \\
		 splice 1 vs. 2 & 93.39 & 92.44 \\
	    USPS 3 vs. 5 & 95.57 & 95.86 \\
	    USPS 5 vs. 8 & 97.71 & 98 \\
	    USPS 7 vs. 9 & 97.57 & 97.43 \\ \hline
	    \end{tabular}
		\caption{Summary of the results: GURU and SVM.}
		\label{tab:binResults}
	\end{center}
\end{table}

\begin{figure}[tbp]
  \centering
    \includegraphics[scale=0.6]{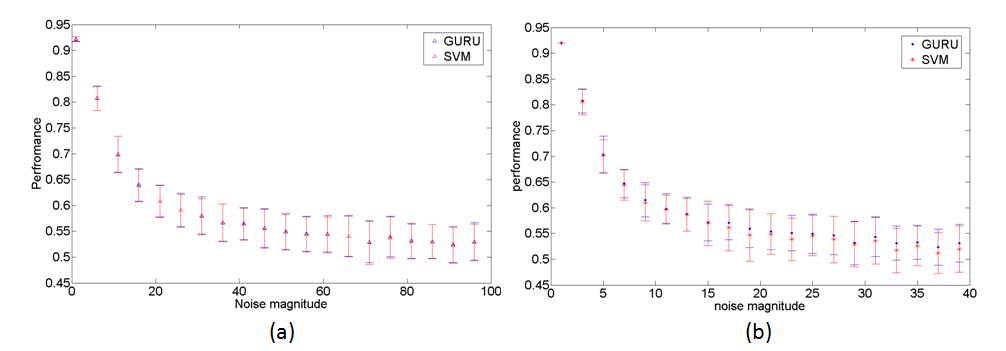}
  \caption{Classifiers' performance on a noised cross-validation and testing sets. The $x$-axis indicates the magnitude of the noise (noise distributes as $U(-x,x)$). The experiment was repeated 50 times. (a)-(b) represent the respective toy problems.}
	\label{fig:linNoise}
\end{figure}

On the toy databases (a)-(b), the performance of GURU is identical to that of SVM. We have tested the learned classifiers' resistance to Noise, by adding uniformly distributed random noise to both cross-validation and test sets. The results are presented in \figref{fig:linNoise}. Observe that the resistance of GURU slightly outperforms that of SVM. Nontheless, this result gives an experimental support to the theoretical result in \cite{XuCaMa09}, where it was shown that the ordinary SVM formulation is equivalent to a robust formulation, in which the adversary is capable of displacing the data samples.

On the Ionosphere database, GURU significantly outperforms SVM. The samples of this database consist of radar reading. Thus, GURU's performance may be understood by the noisy nature of the samples. This finding supports the intuition that GURU perfoms well in noisy setups.

\begin{figure}[tbp]
  \centering
    \includegraphics[scale=0.5]{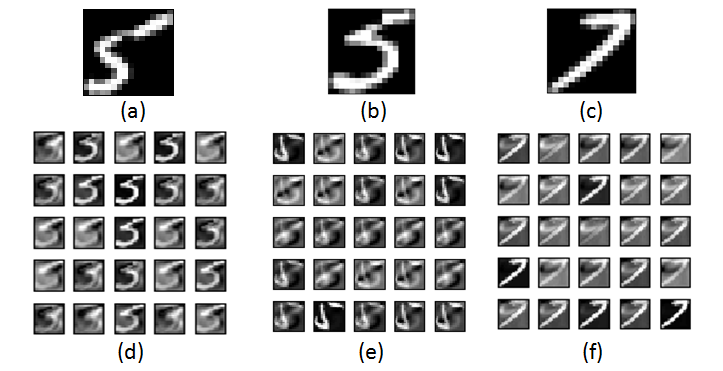}
  \caption{The GURU adversary adds noise perpendicularly to the separating hyperplane. Note that some samples are even more clear than the original, whereas others look like the opponent digit. A bunch of samples are a superposition of both. (a)-(c) are the original digits. (d)-(f) are 25 noisy samples.}
	\label{fig:USPS_adver_action}
\end{figure}

On USPS, the performance of GURU is pretty similar to that of SVM. Since the samples can be easily visualized as images, it is convenient to examine the adversarial action in this case. Consider \figref{fig:USPS_adver_action}. The GURU adversary is symmetric, in the sense that it may move the samples either closer or further from the separating hyperplane. Hence, some digits look even more clear that the original ones, whereas others look as the opponent digit.

\begin{figure}[tbp]
  \centering
    \includegraphics[scale=0.6]{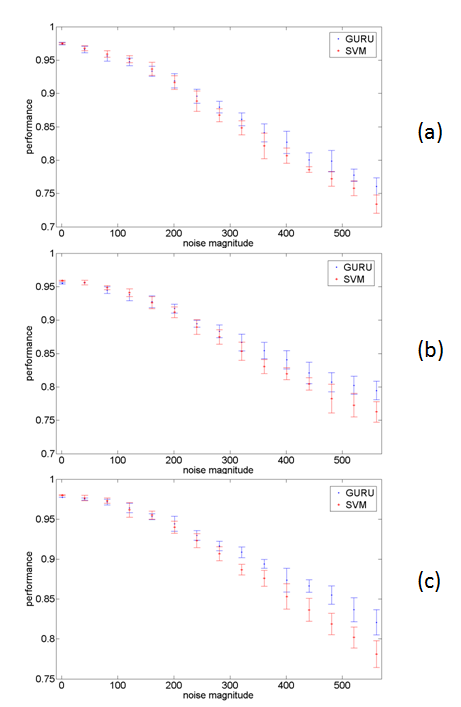}
  \caption{Classifiers' performance on a noised cross-validation and testing sets. The $x$-axis indicates the manitude of the noise (noise distributes as $U(-x,x)$). The experiment was repeated 10 times. (a) 3 vs 5, (b) 5 vs 8, (c) 7 vs 9.}
	\label{fig:USPSnoise}
\end{figure}

In addition, on the USPS dataset, GURU has demonstrated a significantly better resistance to noise than SVM (see \figref{fig:USPSnoise}).

\chapter{Dual formulation}
\label{section-dual}

In this chapter we derive a dual probelm for the learning task at hand. We do not use the dual as a means to solve the primal problem, since the primal optimization works well. Rather, we use it to gain a better understanding of the problem. In the course of the derivation we use the notion of conjugate functions. We will show that the dual problem itself specifies the classifier up to a scailing factor. Thus, we devise a method to extract the norm using the available information.
It is interesting to observe that throughout the derivation of the dual, the smooth function $f$ plays a specific and distinguished role. Thus, the entire procedure may be applied as is for other smooth convex function, by only calculating their conjugate dual. We demonstrate this principle in \secref{sec:genFW}, where we also discuss the relation between the primal loss and the dual formulation.

\section{Mathematical Derivation}
This section is rather technical, and goes through the derivation of the dual. We start with the perspective representation of $\lrh$, and introduce copule of auxilliary variables. Using these variables, the Lagrangian takes a form that we are able to analyze. \thmref{thm:g_th_f} encapsulates the effect of $f$, in such a manner that other loss functions can be plugged into the derivation rather easily. 

The main result of this section is that the dual form of \eqref{eq:forGURU} is
 \be
\begin{array}{ll}
\max &  \sum_m \alpha_m  \\
\mbox{s.t.} &    \|\sum_m \alpha_m y^m \xx^m\| \leq \sigma \sum_m {1\over{\sqrt{2\pi}}}\exp\left({-{\inverf^2(\alpha_m)\over 2}}\right) \\
		& \alv \geq 0 
\end{array}
 \ee
In the following paragraphs we will go through the details.

The optimization task \eqref{eq:forGURU} may be written as
\be
\min_{\ww} { \sigma\|\ww\| \sum_m f\left(\frac{1-y^m\ww^T\xx^m}{\sigma\|\ww\|}\right)}
\ee

We introduce the auxilliary varibles $z_m$, and constrain them with $1-y^m\ww^T\xx\leq z_m$. Note that $f(z)$ is monotone increasing in $z$. Thus, at optimality $z_m=1-y^m\ww^T\xx$. In addition, we introduce the variable $r$ and constrain it with $\sigma\|\ww\|\ \leq r$. At the optimum $r=\sigma\|\ww\|$, since $rf({z\over r})$ is monotone increasing in $r$. Altogether we get the following optimization task
\be
\begin{array}{ll}
\min & r \sum_m f\left(\frac{z_m}{r}\right) \\
\mbox{s.t.} & \sigma\|\ww\|\ \leq r\\
		& 1-y^m\ww^T\xx^m \leq z_m \\
		& r\geq 0
\end{array}		
\ee
where the optimization variables are $\ww,z_1,\ldots,z_n,r$.

 The objective is convex according to \lemref{lem:persIsCvx}, and the constraint on $\ww$ is a second order cone.
 
 To find the dual, write the Lagrangian:
 \bea
 \lag(\ww,r,\zz,\alv,\lambda) &=& r \sum_m  f\left(\frac{z_m}{r}\right)  + \lambda\left[\sigma\|\ww\| - r\right] + \sum_m \alpha_m\left[1-y^m\ww^T\xx^m- z_m\right] -\mu r\\
 \eea
where $\lambda\alpha_m,\mu\geq 0$ are the Lagrange multipliers. For later convenience we add a set of variables $r_m$ and force them all to equal $r$. So the new Lagrangian is:
\bea
 \lag(\ww,r,\zz,\alv,\lambda,\deltav) &=&  \sum_m r_m f\left(\frac{z_m}{r_m}\right)  + \lambda\left[\sigma\|\ww\| - r\right] \\
&+& \sum_m \alpha_m\left[1-y^m\ww^T\xx^m- z_m\right] -\mu r - \sum_m \delta_m \left[ r_m - r\right] \\
\eea
where $\delta_m\geq 0$.

Recall that we have defined $g(a,b) = af\left({b\over{a}}\right)$. Using this notion we get the following task
\bea
\label{eq:forNorm}
\min_{\ww,r,\zz} \lag(\ww,r,\zz,\alv,\lambda,\deltav) &=& \min_{\ww,r}  \sum_m \min_{z_m,r_m}\left[ g(r_m,z_m) - \alpha_m z_m - \delta_m r_m \right]  \\
 &&+ \lambda\left[\sigma\|\ww\| - r\right] + \sum_m \alpha_m\left[1-y^m\ww^T\xx^m\right] -\mu r + r \sum_m \delta_m \\
		&=& \min_{\ww,r} \sum_m g^*(\alpha_m;\delta_m)  + \lambda\left[\sigma\|\ww\| - r\right] \\
&+& \sum_m \alpha_m\left[1-y^m\ww^T\xx^m\right] - \mu r + r\sum_m \delta_m
\eea
where $g^*$ is by definition the conjugate function of $g$ (for details see \cite{BoydOpt} $3.3$, e.g). Deriving the Lagrangian w.r.t. $\ww$ gives:
 \be
\label{eq:dualRep}
\sigma \lambda {\ww\over \|\ww\|} = \sum_m \alpha_m y^m \xx^m
 \ee
Taking the norm of both sides of the equation yields
 \be
\sigma \lambda(\alv) = \|\sum_m \alpha_m y^m \xx^m\|
 \label{eq:lambda_sol}
 \ee
Substituting this back into the objective, the terms with $\ww$ cancel out and we have:
 \be
 \min_r \sum_m g^*(\alpha_m;\delta_m) - r\lambda(\alv) - \mu r + r\sum_m \delta_m
 \ee
 This is linear in $r$, thus deriving w.r.t. $r$ yields a constraint $\sum_m \delta_m = \lambda(\alv) + \mu$. Since $\mu\geq 0$, the equality constraint might be relaxed to $\sum_m \delta_m \geq \lambda(\alv)$, and we end up with the following formulation
 \be
\begin{array}{ll}
\max &  \sum_m \alpha_m + \sum_m g^*(\alpha_m;\delta_m) \\
\mbox{s.t.} & \sum_m \delta_m \geq \lambda(\alv) \\
		& \alv \geq 0 
\end{array}
 \ee
Or:
 \be
\begin{array}{ll}
\max &  \sum_m \alpha_m + \sum_m g^*(\alpha_m;\delta_m) \\
\mbox{s.t.} &    \|\sum_m \alpha_m y^m \xx^m\| \leq \sigma \sum_m \delta_m \\
		& \alv \geq 0 
\end{array}
\label{eq:dual_take_one}
 \ee
  The overall problem has a concave objective (since it's a conjugate dual of a convex function) and second order cone constraints. In what follows we work
  out the form of the conjugate dual $g^*$.
 
Denote by $f^*(\alpha)$ the conjugate function of $f$ (it is concave). The next theorem specifies the conjugate $g^*$ in terms of $f^*$:
 
\thm
\label{thm:g_th_f}
The conjugate dual of $g(a,b)$ is
\be
\label{eq:gConj}
 g^*(\alpha,\delta)  = 
 \left\{
 \begin{array}{ll}
 0 & f^*(\alpha)\geq \delta \\
 -\infty & {\mbox otherwise} 
 \end{array}
 \right.
 \ee
\ethm
\prf
We must calculate
 \be
 g^*(\alpha;\delta) = \min_{x,t}\left(tf({x \over t}) - \alpha x - \delta t\right)
 \ee
 
To prove, we change from variables $x,t$ to a variables $z = x/t,t$:
 \be
  \min_{t\geq 0,z} tf(z) - \alpha zt - \delta t = \min_{t\geq 0,z} t(f(z) - \alpha z - \delta)
 \label{eq:dual_xz_min}
 \ee
 For the first case, assume that $f^*(\alpha)\geq \delta$, which implies that for all $z$:
 \be
 f(z) - \alpha z \geq \delta
 \ee
 Then in \eqref{eq:dual_xz_min} the minimization is always of the product of $t\geq 0$ and some non-negative number. Hence it is always greater than zero, 
 and zero can be attained at the limit $t\to 0$. 
 
 On the other hand if  $f^*(\alpha)< \delta$, we will show that there exists a pair $t,z$ that achieves a value $-\infty$: 
 Since $f^*(\alpha)< \delta$ there exists a $z$ for which 
 \be
 f(z) - \alpha z - \delta <0
 \ee
 If we take $t\to \infty$ and this $z$ we get a value of $-\infty$. 
\eprf 

In order the complete the derivation of the dual formulation, we should compute the conjugate dual $f^*$. The following lemma gives the desired result
\lem
\label{lem:fConj}
The conjugate dual of $f$ is
\be
f^*(\alpha) = {1\over{\sqrt{2\pi}}}\exp\left({-{\inverf^2(\alpha)\over 2}}\right) \nonumber
\ee
\elem
\prf
  Recall that:
 \be
f(z) = z\erf(z)+{1\over{\sqrt{2\pi}}}e^{-{z^2\over 2}}
\ee
and that its first derivative is 
\be
{df\over{dz}} = \erf(z) \nonumber
\ee
(see \eqref{eq:fDeriv}).
By \thmref{thm:fProp}, $f$ is convex, thus we compute $f$'s conjugate dual:
\be
f^*(\alpha) = \min_{z} f(z)-\alpha z
\ee
The minimum satisfies:
\bea
f'(z) &=& \alpha \\
\erf(z) &=& \alpha \\
z &=& \inverf(\alpha)
\eea
where $\inverf$ is the inverse function of $\erf$.
We plug this equality into the objective and conclude
\bea
f^*(\alpha) &=& f(\inverf(\alpha)) - \alpha \inverf(\alpha) \\
		&=&  \inverf(\alpha)\alpha + {1\over{\sqrt{2\pi}}}\exp\left({-{\inverf(\alpha)^2\over 2}}\right)- \alpha \inverf(\alpha) \\
		&=& {1\over{\sqrt{2\pi}}}\exp{\left({-{\inverf^2(\alpha)\over 2}}\right)}
\eea
It can be easily verified that $f^*$ is concave, as expected from the theory.
Note that from the derivation above it follows that $\alpha_M\leq 1$.
\eprf

Taking the dual problem in \eqref{eq:dual_take_one} and plugging in the conjugate duals derived above, we get:
 \be
\begin{array}{ll}
\max &  \sum_m \alpha_m  \\
\mbox{s.t.} &    \|\sum_m \alpha_m y^m \xx^m\| \leq \sigma \sum_m \delta_m \\
		& {1\over{\sqrt{2\pi}}}\exp\left({-{\inverf^2(\alpha_m)\over 2}}\right) \geq \delta_m \\
		& \alv \geq 0 
\end{array}
\label{eq:dual_take_two}
 \ee
Consider the following problem:
 \be
\begin{array}{ll}
\max &  \sum_m \alpha_m  \\
\mbox{s.t.} &    \|\sum_m \alpha_m y^m \xx^m\| \leq \sigma \sum_m {1\over{\sqrt{2\pi}}}\exp\left({-{\inverf^2(\alpha_m)\over 2}}\right) \\
		& \alv \geq 0 
\end{array}
\label{eq:dual_take_two_reduced}
 \ee

The following proposition asserts that both of the formulations above are equivalent.
\pro
\eqref{eq:dual_take_two} and \eqref{eq:dual_take_two_reduced} are equivalent.
\epro
\prf
Denote $\mathcal{C}_1$ the feasible region of \eqref{eq:dual_take_two}, and $\mathcal{C}_2$ the feasible region of \eqref{eq:dual_take_two_reduced}.
	Let $\alv\in\mathcal{C}_1$. Then trivially we have $\alv\in\mathcal{C}_2$.
 On the other hand, let $\alv\in\mathcal{C}_2$. Denote 
$\delta_m = {1\over{\sqrt{2\pi}}}\exp(-{1\over{2}}\inverf(\alpha_m)^2)$. It is easy to verify that this selection corresponds to a feasible point for \eqref{eq:dual_take_two} (i.e. $\in\mathcal{C}_1$) with the same objective value.
\eprf

As we have seen, the optimization problem we analyze in this work is a relative of the SVM problem. It is interesting to examine what happens when considering the duals.
Consider the SVM formulation
 \be
\begin{array}{ll}
\min_{\ww} & \frac{\lambda}{2}\|\ww\|^2+ \sum_{m=1}^M{\xi_m}  \\
\mbox{s.t.} & \forall  m\in\{1,2,\ldots,M\}: \xi_m\geq 1-y^m\ww^T\xx^m \\
		& \xi_m \geq 0
\end{array}
 \ee
Its dual is
 \be
\begin{array}{ll}
\min_{\alv} & \sum_{m=1}^M{\alpha_m}-\frac{1}{2}\sum_{m,n=1}^M{\alpha_m\alpha_n y^m y^n (\xx^m)^T\xx^n} \\
\mbox{s.t.} & \forall  m\in\{1,2,\ldots,M\}:0\leq\alpha_m\leq \frac{1}{\lambda}
\end{array}
 \ee
This dual form shares some properties with the dual form of GURU. For example, notice that in both cases one tries to maximize the sum of the dual variables $\alpha_m$. Another issue is that of the norm minimization. The SVM dual explicitly minimizes the norm of the classifier. In our dual, however, the situation is rather implicit: there exist a bound on the norm of the classifier. Without going into the details, we mention that moving a constraint into the objective or vice versa is possible in the context of Lagrangian duality. At last, notice that in spite of the fact that $\sigma$ and $\lambda$ play similar roles, increasing $\lambda$ results in srinking the feasible region of the SVM duak, whereas in our problem, increasing $\sigma$ expands the feasible region.

 The last issue we discuss in this section is the norm of the optimal classifier. Note that by solving the dual formulation, one can only get the optimal classifier up to a scailing factor. Of course, it is essential to know the norm exactly in order to be able to use the classifier. This goal can be achieved using the following theorem:
\thm
\label{eq:restoreNorm}
The norm of the optimal classifier is
\be
\|\ww^*\|={1\over{\inverf({\alpha_m}^*)+y^m{({\hat{\ww}}^*)}^T\xx^m}}
\ee
for every $m$, where ${\hat{\ww}}^*$ is the normalized optimal classifier.
\ethm

\prf
\eqref{eq:forNorm} may be written as
\bea\min_{\ww,r,\zz} \lag(\ww,r,\zz,\alv,\lambda,\deltav) &=& \min_{\ww,r}  \sum_m \min_{z_m,r_m}\left[ r_m f\left({z_m\over{r_m}}\right) - \alpha_m z_m - \delta_m r_m \right]  \\
 &&+ \lambda\left[\sigma\|\ww\| - r\right] + \sum_m \alpha_m\left[1-y^m\ww^T\xx^m\right] -\mu r + r \sum_m \delta_m \\
	&=& \min_{\ww,r}  \sum_m \min_{r_m}\left[r_m\min_{z_m} \left[ f\left({z_m\over{r_m}}\right) - \alpha_m {z_m\over{r_m}}\right] - \delta_m r_m\right]\\
 &&+ \lambda\left[\sigma\|\ww\| - r\right] + \sum_m \alpha_m\left[1-y^m\ww^T\xx^m\right] -\mu r + r \sum_m \delta_m \\
\eea
since $r_m\geq 0$. We define $q_m={z_m\over{r_m}}$. Since the equation above depends on $z_m$ only via $q_m$, we get

\bea\min_{\ww,r,\zz} \lag(\ww,r,\zz,\alv,\lambda,\deltav) &=& \min_{\ww,r}   \sum_m \min_{r_m}\left[r_m\min_{q_m} \left[ f\left(q_m\right) - \alpha_m q_m\right] - \delta_m r_m\right]\\
 &&+ \lambda\left[\sigma\|\ww\| - r\right] + \sum_m \alpha_m\left[1-y^m\ww^T\xx^m\right] -\mu r + r \sum_m \delta_m \\
\eea
If when substituting the dual optimum in the Lagrangian, there exists a unique primal feasible solution, then it must be primal optimal (see \cite{BoydOpt}, $5.5.5$ for details). Thus, at the optimum $q_m^* = \min_{q_m} \left[ f\left(q_m\right) - \alpha_m q_m\right]$. According to the proof of \lemref{lem:fConj} it holds that $q_m^* = \inverf(\alpha_m)$. By exploiting the monotony properties of the problem (that were presented in the beginning of the section), we conclude that
\be
\label{eq:forBound}
{{1-y^m\ww^T\xx^m}\over{\sigma\|\ww\|}}=\inverf(\alpha_m)
\ee
The desired result follows from basic algebraic operations.
\eprf

Note that the values of the optimal $\alpha$'s are known, as well as the normalized vector $\hat{\ww}^*={{\ww^*}\over{\|\ww^*\|}}$. Thus, we can compute the optimal norm.

\begin{figure}[tbp]
  \centering
    \includegraphics[scale=0.6]{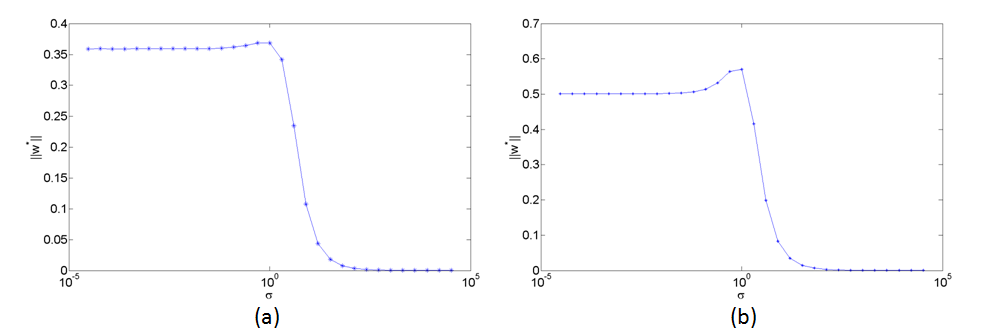}
  \caption{Norm of the optimal classifiers trained for the toy problems of \secref{sec:bExp}, for various $\sigma$ values. (a)-(b) represent the respective toy problems.}
	\label{fig:linNorm}
\end{figure}

It is possible that the norm of the optimal classifier is bounded (as a function of $\sigma$). Although we couldn't prove this result, we conjecture that such a result might stem from a strong duality argument:
\be
\|\alv^*\|_1 = \sum_m\lrh(\xx^m,y^m;\ww^*,\sigma^2)
\ee
By plugging \eqref{eq:forBound} into the previous equality, we obtain
\be
\|\ww\|={{\|\alv^*\|_1}\over{\sum_m{f(\inverf\left(\alpha_m\right))}}}
\ee
A better understanding of the constraints on $\alv$ may help bounding the RHS of the equation.
We have plotted the norm of the optimal classifiers for the toy problems of \chapref{section-GURU} (refer to \secref{sec:bExp} for more details). The results are shown in \figref{fig:linNorm} and clearly support this conjecture.

\section{A general framework}
\label{sec:genFW}
The dual form we have derived sheds some light on the structure of the problem. In this section we discuss the relation between the loss function $f$ and the norm constraint that appears in the dual. We claim that there is a correspondence between approximations of $f$ and relaxations of the dual problem. More specifically, approximations of the loss function culminates in approximations of the feasible region of the dual problem.

\begin{figure}[h!]
  \centering
    \includegraphics[scale=0.4]{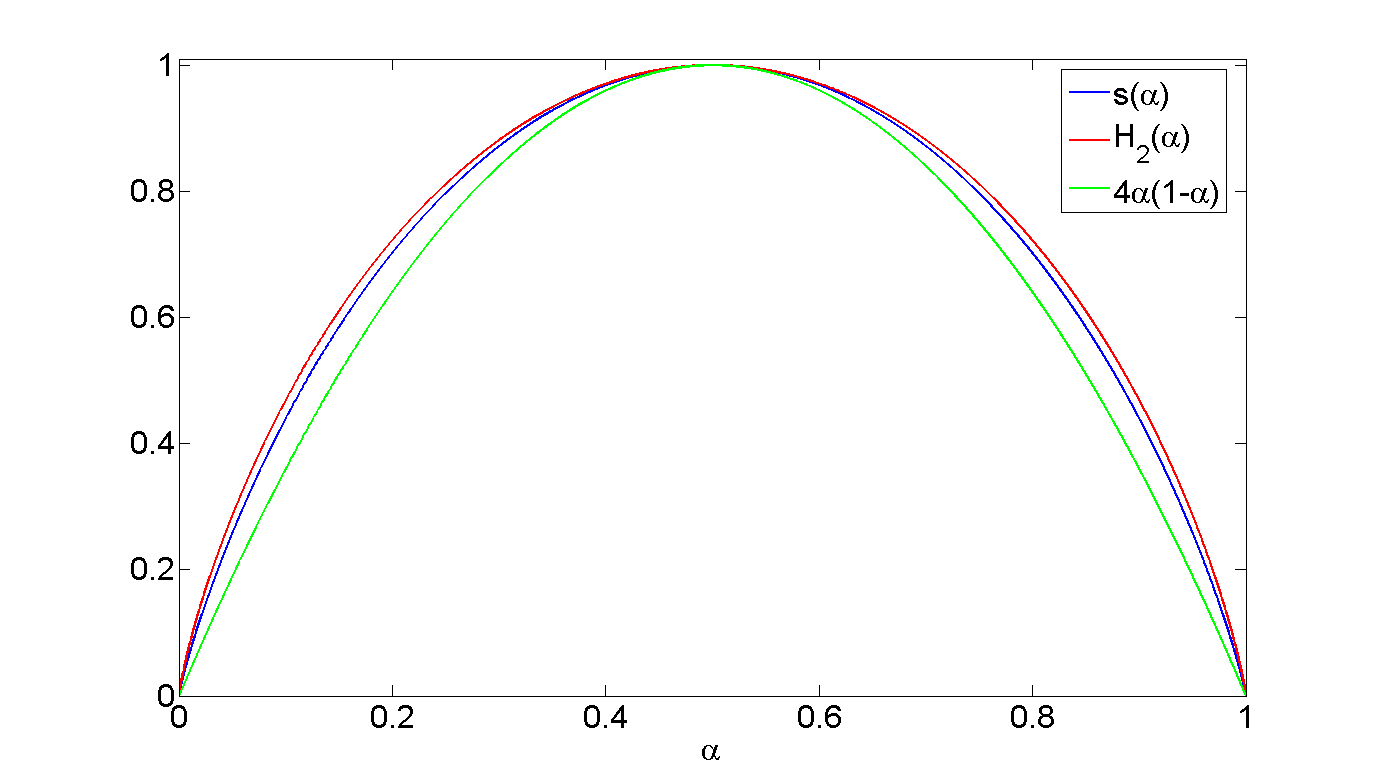}
  \caption{The dual constraint may be approximated using elementary functions.}
\label{fig:consApprox}
\end{figure}

The norm constraint in the dual is a core component of the optimization. We denote by
\be
\label{eq:originalCon}
s(\alpha) = \exp\left(-\frac{\inverf(\alpha)^2}{2}\right)
\ee
the function under summation. It is complicated to handle and understand $s(\alpha)$, thus it is appealing to approximate it using elementary functions. Two such approximations are
\be
\label{eq:approxes}
\tilde{s}_1(\alpha) = H_2(\alpha) = -\alpha\log_2(\alpha)-(1-\alpha)\log_2(1-\alpha) \nonumber
\ee
\be
\tilde{s}_2(\alpha) = 4\alpha(1-\alpha)
\ee
(see \figref{fig:consApprox}).

Note that in the previous section we only used $f$ as a means to express $g^*$ (\eqref{eq:gConj}). Thus, if one replaces $f$ with some alternative convex loss function $\tilde{f}$, the derivation of the dual will remain correct. Of course, the dual norm constraint will be affected by this change. \newline
In order to understand the nature of the approximations in \eqref{eq:approxes}, it is necessary to explore the respective dual conjugates.
\lem
Let $\tilde{f}_2(z)=\log_2(1+2^z)$. Then its conjugate dual is
\be
\tilde{f}_2^*(\alpha)=-\alpha\log_2(\alpha)-(1-\alpha)\log_2(1-\alpha)
\ee
\elem
\prf
We compute $\tilde{f}_2$'s conjugate dual:
\be
\tilde{f}_2^*(\alpha) = \min_{z} \tilde{f}_2(z)-\alpha z
\ee
The minimum satisfies:
\bea
\tilde{f}_2'(z) &=& \alpha \\
\frac{2^z}{2^z+1} &=& \alpha \\
2^z &=& \frac{\alpha}{1-\alpha} \\
z &=& \log_2\left(\frac{\alpha}{1-\alpha}\right)
\eea
We plug this equality into the objective and conclude
\bea
\tilde{f}_2^*(\alpha) &=& \log_2{\left(1+\frac{\alpha}{1-\alpha}\right)}-\alpha\log_2\frac{\alpha}{1-\alpha} \\
&=& -\alpha\log_2(\alpha)-(1-\alpha)\log_2(1-\alpha)
\eea
as claimed. As in the case of our Gaussian robust loss, we have $\alpha\leq 1$.
\eprf

\begin{figure}[h!]
  \centering
    \includegraphics[scale=0.4]{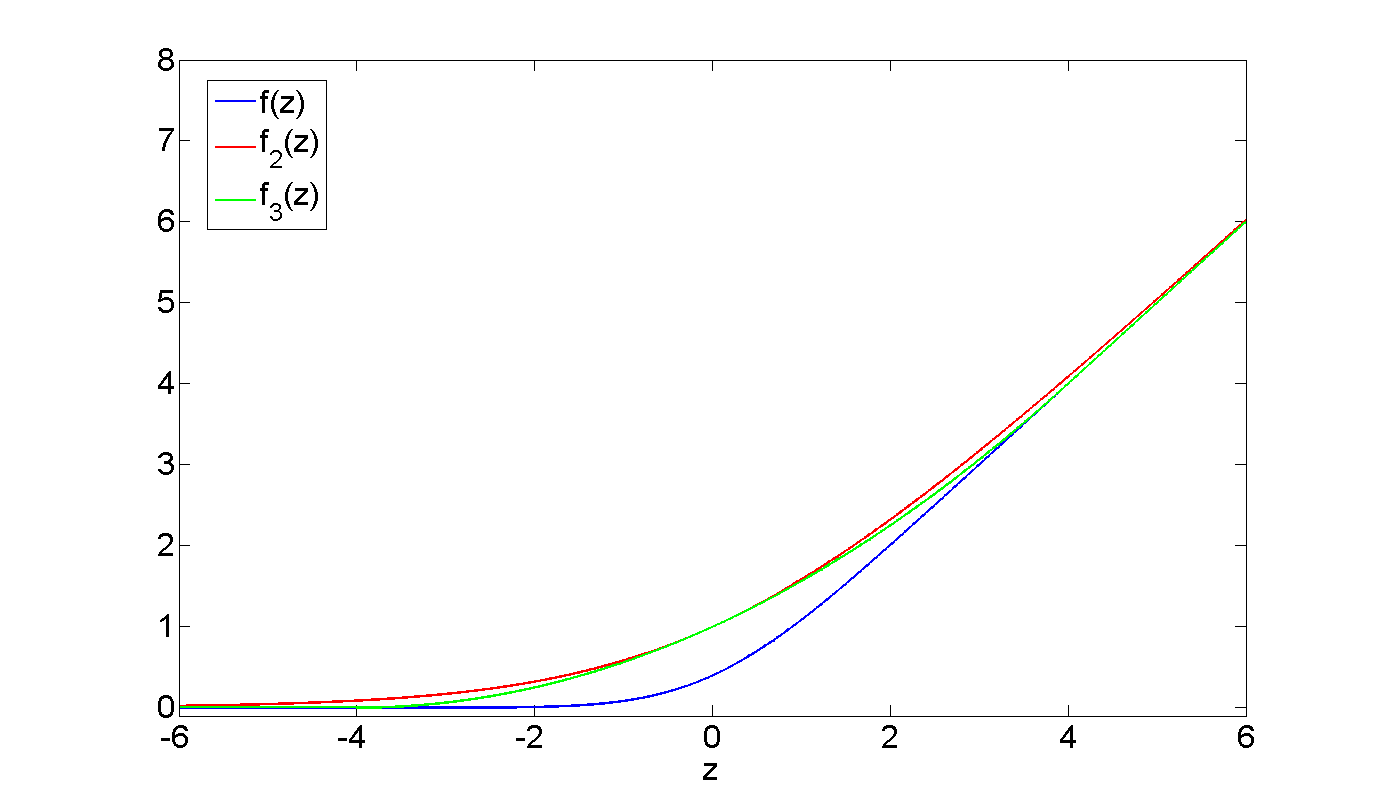}
  \caption{Log loss apears naturally in our framework. In addition, we have demonstrated a means to generate some other loss functions, such as the quaratic loss above.}
\label{fig:diffLosses}
\end{figure}

\lem
Let
\be
\tilde{f}_3(z)=
	\begin{cases}
		0 & \text{if } z<-4 \\
		{{(z+4)^2}\over{16}}& \text{if } -4\leq z \leq 4 \\
		z & \text{if } z>4 \\
  \end{cases}
\ee
Then its conjugate dual is
\be
\tilde{f}_3^*(\alpha)=
\begin{cases}
		4\alpha(1-\alpha) & \text{if } 0\leq\alpha\leq 1 \\
		-\infty & \text{if } \alpha>1 \\
  \end{cases}
\ee
\elem

\prf
It is easy to verify that $\tilde{f}_3$ is smooth. We thus compute $\tilde{f}_3$'s conjugate dual in the following way:
\be
\tilde{f}_3^*(\alpha) = \min_{z} \tilde{f}_3(z)-\alpha z
\ee
Extremum points satisfy:
\bea
\tilde{f}_3'(z) -\alpha &=& 0 \\
	\begin{cases}
		-\alpha & \text{if } z<-4 \\
		{(z+4)\over{8}}-\alpha & \text{if } -4\leq z \leq 4 \\
		1-\alpha & \text{if } z>4 \\
  \end{cases}
&=& 0
\eea
The above equation vanishes at $z=8\alpha-4$. For $0\leq\alpha\leq 1$ we have $-4\leq z\leq 4$, thus we conclude
\bea
\tilde{f}_3^*(\alpha)\Big|_{0\leq\alpha\leq 1} &=& 4\alpha(1-\alpha)
\eea
For $\alpha>1$ we take $z\to\infty$, and $\tilde{f}_3(z)=\Big|_{z>4}(1-\alpha)z\to -\infty$.
Altogether we hace established the desired result.
\eprf \newline

These lemmas shade some light on $\lrh$ and on the structure of our problem. It turns out that the well-known log-loss as well as a quadratic loss that has the same flavour as the Huber loss appear naturally in our framework (see \figref{fig:diffLosses} for a visualization). What we have demonstrated is that there exist a close connection between approximations of the primal loss and relaxations of the dual problem. Specifically, we have that the dual of
\be
\min_{\ww}{\sum_{m=1}^M{\|\ww\|\tilde{f}\left({{1-y^m\ww^T\xx^m}\over{\|\ww\|}}\right)	}
}
\ee
is
 \be
\begin{array}{ll}
\max &  \sum_m \alpha_m  \\
\mbox{s.t.} &    \|\sum_m \alpha_m y^m \xx^m\| \leq \sum_m \tilde{s}(\alpha_m) \\
		& \alv \geq 0 
\end{array}
 \ee

Note, however, that this connection should be further investigated. It should be observed that not every smooth convex primal loss $\tilde{f}$ yields a perspective that is convex in $\ww$. For that to happen, $\tilde{f}$ should satisfy some mathematical properties that are yet to be understood. One example for such a condition is $f(z)\geq z\frac{d\tilde{f}}{dz}(z)$. Under this condition we can use the same reasoning as in the proof of \thmref{thm:actualCvx} and conclude that the primal probem is convex. In this case, we can automaticaly apply the derivation presented in the previous section and deduce the respective dual problem. Another issue that should be better understood is the connection between approximations of $f$ and the robust setup we have begun with. In particular, it is interesting to understand if the logarithmic loss may be interperted as resulting from RO.

\chapter {Introducing Kernels}
\label{section-kernels}
One of the greatest stengths of the theory of support vector machines, is the simple generalization to nonlinear cases. This generalization is carried out via the elegant notion of kernels. An examination of our derivation suggests that one may apply the kernel trick and introduce a means to learn nonlinear classifiers in Gaussian Robust framework. 

In this chapter we will develop a kernelized version of the GURU algorithm. Most of the derivation is straight forward: we begin by giving a representer result. Plugging the new parametrization of the classifer into the framework, we show that our update formulas are perfectly suitable for maintaining this kind of representation. The tricky part stems from the fact that our updates depend directly on the norm of the weights vector. Naive computation of the norm costs $O(M^2)$ operations, which significantly slows down the algorithm. We thus derive a procedure to update the norm in $O(1)$, based on previous computations.

\section {A representer result}
The first step towards kernelization of GURU, is to change our represention of the classifier from a weights vector ($\ww$) to a linear combination of the training samples. The theoretical justification of such an operations is known as a representer result.\newline
The fact that an optimal classifier may be represented as a linear combination of the training sample, stems from the mathematical theory of Hilbert spaces. In our case, as well as in SVM, however, the same result can be derived using far more simple and explicit argumentation. In this section we will show three ways to establish the representer result for the case of GURU. In spite of the fact that we could prove the theroem using abstract argumentation, it is necsssary to develop the technical proof, as it lays the foundations for the derivation of the kerenelized algorithm.

We start by stating a version of the representer theorem:
\thm
\label{thm:whaba}
Let $\mathcal{H}$ be a reproducing kernel Hilbert space with a kernel $\kappa:\mathcal{X}\times\mathcal{X}\to\mathbb{R}$, a symmetric positive semi-definite function on the compact domain. For any function $L:\mathbb{R}^n\to\mathbb{R}$, and any nondecreasing function $\Omega:\mathbb{R}\to\mathbb{R}$. If
\be
{J^*}=\min_{f\in\mathcal{H}}{J(f)} = \min_{f\in\mathcal{H}}{\left\{\Omega\left({\|f\|}_{\mathcal{H}}^2\right) +L\left(f(x_1),f(x_2),\ldots,f(x_n)\right)\right\}} \nonumber
\ee
is well-defined, then there are some $\alpha_1,\alpha_2,\ldots\alpha_n\in\mathbb{R}$, such that
\be
\label{eq:genRep}
f(\cdot)=\sum_{i=1}^n{\alpha_i\kappa(x_i,\cdot)}
\ee
acheives $J(f)=J^*$. Furthermore, if $\Omega$ is increasing, then each minimizer of $J(f)$ can be expressed in the form of \eqref{eq:genRep}.
\ethm
For a proof and more details, see for example \cite{LWK}.

As mentioned, we will discuss three techniques to establis the required result. First, using the structure of the updates that GURU perform. Second, by the derivation of the dual problem presented in \chapref{section-dual}, and third, using the general representer theorem.

\thm
\label{thm:repThm}
There exists a solution of \eqref{eq:forGURU} that takes the form
\be
\ww=\sum_{m=1}^M{\alpha_m y^m\xx^m}
\ee
\ethm

\prf \textbf{Via the structure of GURU} \newline
Recall that the updates in the GURU algorithm are of the form

$$\ww \leftarrow \ww-{\eta\over \sqrt{t}}\left(-y^i\xx^i \erf\left({1-y^i\ww^T\xx^i}\over {\sigma\|\ww\|}\right)+{\sigma\ww\over\sqrt{2\pi}\|\ww\|}\exp\left(-{(1-y^i\ww^T\xx^i)^2\over 2\sigma^2\|\ww\|^2}\right)\right)
$$
It is suggestive to observe that the update formula can be split and written as two successive steps. The first of which is
$$\ww \leftarrow \ww-{\eta\over \sqrt{t}}{\sigma\ww\over\sqrt{2\pi}\|\ww\|}\exp\left(-{(1-y^i\ww^T\xx^i)^2\over 2\sigma^2\|\ww\|^2}\right)$$
followed by

\be
\label{eq:scSpec}
\ww \leftarrow \ww+{\eta\over \sqrt{t}}y^i\xx^i \erf\left({1-y^i\ww^T\xx^i}\over {\sigma\|\ww\|}\right)
\ee

The first step is nothing else then a rescailing of the weights vector
\be
\label{eq:compGamma}
\ww = \gamma\ww,\;\;\gamma = 1-{\eta\over \sqrt{t}}{\sigma\over\sqrt{2\pi}\|\ww\|}\exp\left(-{(1-y^i\ww^T\xx^i)^2\over 2\sigma^2\|\ww\|^2}\right)
\ee

Recall that GURU initializes the weight vector as $\ww=0$, which clearly can be represented as
\be
\boldsymbol{0}=\sum_{m=1}^M{0 y^m\xx^m}
\ee
We thus assume that the desired representation exists, and proceed by induction. By plugging the representation into the previous equations, we get
\be
\sum_{m=1}^M{\alpha_m^{new} y^m\xx^m} = \sum_{m=1}^M{\alpha_m y^m\xx^m} + \gamma\sum_{m=1}^M{\alpha_m y^m\xx^m} \nonumber
\ee
i.e. for all $m$
\be
\label{eq:upRSC}
\alpha_m^{new} = (1+\gamma)\alpha_m
\ee
where $\alpha_m^{new}$ is the result of thee respective update. The second step in the update formula (\eqref{eq:scSpec}), may be written as
\be
\label{eq:compMu}
\sum_{m=1}^M{\alpha_m^{new} y^m\xx^m}=\sum_{m=1}^M{\alpha_m y^m\xx^m} + \mu_i y^i\xx^i,\;\;\mu_i={\eta\over \sqrt{t}}y^i\xx^i \erf\left({1-y^i\ww^T\xx^i}\over {\sigma\|\ww\|}\right) \nonumber
\ee
i.e.
\be
\label{eq:upOO}
 \alpha_m^{new} =
  \begin{cases}
   \alpha_m & \text{if } m\neq i  \\
   \alpha_i + \mu_i & \text{if } m = i
  \end{cases}
\ee

Combining both steps, we end up with the following update rule:
\be
\label{eq:upTot}
 \alpha_m^{t+1} =
  \begin{cases}
   \gamma\alpha_m^t & \text{if } m\neq i  \\
   \gamma\alpha_i^t + \mu_i & \text{if } m = i
  \end{cases}
\ee
Since GURU is guranteed to converge to the optimum, by taking $t\to\infty$ we establish the desired result.
\eprf

\prf \textbf{Via the dual formulation} \newline
We have already seen (\eqref{eq:dualRep}) that
\be
\sigma \lambda {\ww\over \|\ww\|} = \sum_m \alpha_m y^m \xx^m\nonumber
\ee
By defining $\tilde{\alpha}_m = {\|\ww\|\over{\sigma\lambda}}\alpha_m$ and plugging it into the previous equality, we conclude that
\be
\ww = \sum_m \tilde{\alpha}_m y^m \xx^m\nonumber
\ee
as required.
\eprf

\prf \textbf{Via the general representer theorem}\newline
Set $\Omega\equiv 0$, $L(\left(f(x_1),f(x_2),\ldots,f(x_n)\right))=\sum_{i=1}^n{f(x_i)}$, $f=\lrh$. and let $\kappa$ be the linear kernel $\kappa(x_1,x_2)=x_1^T x_2$. The desired result stems immidiately from \thmref{thm:whaba}.
\eprf

\section{KEN-GURU: A primal kernelized version of GURU}
In the pevious section we have established a representer result for GURU. The next step in the derivation is to work the components of the algorithm, so the only dpendence on the data samples and on the classifier would be via dot products. That being the case, we can apply the kernel trick, namely to replace each dot product ${\left(\xx^m\right)}^T\xx^n$ with the kernel entry $\kappa(\xx^m,\xx^n)$ (for details see, for example,  \cite{Aizerman67theoretical,LWK}). We start by expanding the quantities that appear in the update formula in terms of $\alpha_m$'s. Then, we introduce a method to update the value of the norm variable in a computationally cheap way.
We conclude the section by putting the results together, and prsenting the KEN-GURU (\textbf{KE}r\textbf{N}elized \textbf{G}a\textbf{U}ssian \textbf{R}ob\textbf{U}st) algorithm.

In order to compute $\gamma$ and $\mu_i$ of \eqref{eq:compGamma} and \eqref{eq:compMu}, one must know the values of $\ww^T\xx^i$ and $\|\ww\|$. Let us expand the first quantity
\bea
\ww^T\xx^i &=& \left(\sum_{m=1}^M{\alpha_m y^m\xx^m}\right)^T\xx^i \nonumber \\
&=& \sum_{m=1}^M{\alpha_m y^m{\left(\xx^m\right)}^T\xx^i} \nonumber \\
&=& \sum_{m=1}^M{\alpha_m y^m K_{mi}}
\eea
The norm might be computed as
\bea
\|\ww\|^2 = \ww^T\ww &=& \left(\sum_{m=1}^M{\alpha_m y^m\xx^m}\right)^T\sum_{n=1}^M{\alpha_n y^n\xx^n}\nonumber \\
&=& \sum_{m=1}^M{\sum_{n=1}^M{\alpha_m\alpha_n y^m y^n{\left(\xx^m\right)}^T\xx^n}} \nonumber \\
&=& \sum_{m=1}^M{\sum_{n=1}^M{\alpha_m\alpha_n y^m y^n K_{mn}}}
\eea

Note that the Gram matrix $K$ may be precomputed and cached (total cost of $O(M^2)$). Thus, $\ww^T\xx^i$ can be computed in $O(M)$, and $\|\ww\|$ in $O(M^2)$. As both of these values should be computed for each update, the cost of the norm computation is extremely expensive. Instead of computing the norm each time from scratch, it is possible to use its previous value. The updated norm may be computed as

\bea
\|\ww\|_{t+1}^2 &=& \sum_{m=1}^M{\sum_{n=1}^M{\alpha_m^{t+1}\alpha_n^{t+1} y^m y^n K_{mn}}} \nonumber \\
&=& \sum_{m=1}^M{\left[\sum_{n\neq i}{\alpha_m^{t+1}\alpha_n^{t+1} y^m y^n K_{mn}}+\alpha_m^{t+1}\alpha_i^{t+1} y^m y^i K_{mi}\right]} \nonumber \\
&=& \sum_{m=1}^M{\sum_{n\neq i}{\alpha_m^{t+1}\alpha_n^{t+1} y^m y^n K_{mn}}}+\sum_{m=1}^M{\alpha_m^{t+1}\alpha_i^{t+1} y^m y^i K_{mi}} \nonumber \\
&=& \sum_{n\neq i}{\left[\sum_{m\neq i}{\alpha_m^{t+1}\alpha_n^{t+1} y^m y^n K_{mn}} + \alpha_i^{t+1}\alpha_n^{t+1} y^i y^n K_{in}\right]} \nonumber \\
	&\qquad& + \sum_{m=1}^M{\alpha_m^{t+1}\alpha_i^{t+1} y^m y^i K_{mi}} \nonumber \\
&=& \sum_{n\neq i}{\sum_{m\neq i}{\alpha_m^{t+1}\alpha_n^{t+1} y^m y^n K_{mn}}} + \sum_{n\neq i}{\alpha_i^{t+1}\alpha_n^{t+1} y^i y^n K_{in}} \nonumber \\
	&\qquad& + \sum_{m\neq i}{\alpha_m^{t+1}\alpha_i^{t+1} y^m y^i K_{mi}} + \alpha_i^{t+1}\alpha_i^{t+1} y^i y^i K_{ii} \nonumber \\
&=& \sum_{n\neq i}{\sum_{m\neq i}{\alpha_m^{t+1}\alpha_n^{t+1} y^m y^n K_{mn}}} + 2\sum_{m\neq i}{\alpha_m^{t+1}\alpha_i^{t+1} y^m y^i K_{mi}} \nonumber \\
	&\qquad& + \alpha_i^{t+1}\alpha_i^{t+1} y^i y^i K_{ii} \nonumber \\
\eea

By plugging \eqref{eq:upTot} we get

\bea
\label{eq:compNu}
\|\ww\|_{t+1}^2 &=& \gamma^2\sum_{n\neq i}{\sum_{m\neq i}{\alpha_m^t\alpha_n^t y^m y^n K_{mn}}} + 2\gamma\sum_{m\neq i}{\alpha_m^t(\gamma\alpha_i^t+\mu_i) y^m y^i K_{in}} \nonumber \\
	&\qquad& + (\gamma\alpha_i^t+\mu_i)^2 K_{ii} \nonumber \\
&=& \gamma^2\|\ww\|_t^2 + 2\gamma\mu_i y^i\sum_{m=1}^M{\alpha_m^t y^m K_{mi}} + \mu_i^2 K_{ii} \nonumber \\
&=& \gamma^2\|\ww\|_t^2 + 2\gamma\mu_i y^i\ww^T\xx^i + \mu_i^2 K_{ii}
\eea
where $\ww^T\xx^i$ is computed regardless of $\|\ww\|^2$. Thus, the value of the norm can be maintained in $O(1)$.

In may be easily observed that the data samples $\xx^m$ participate in the computations of the update only via the Gram matrix $K$. Thus, we can apply the kernel trick, and use
\be
K_{ij}=\kappa(\xx^i,\xx^j)
\ee
for any Mercer Kernel $\kappa$. Based on the results established in the previous sections, we may translate GURU into a kerenlized version, named KEN-GURU. \newline
We intoduce an auxilliary variable $\zeta$, that holds the value of the product $\kappa(\ww,\xx^i)$ and is evaluated by
\be
\label{eq:compZeta}
\zeta_{t+1} = \sum_{m=1}^M{\alpha_m^t y^m K(\xx^m,\xx^i)}
\ee
According to \eqref{eq:compGamma}, \eqref{eq:compMu} and \eqref{eq:compNu} we introduce the following update formulas
\be
\label{eq:formGamma}
\gamma_{t+1} = 1-{\eta\over \sqrt{t}}{\sigma\over\sqrt{2\pi}\nu_t}\exp\left(-{(1-y^i\zeta_{t+1})^2\over 2\sigma^2\nu_t^2}\right)
\ee

\be
\label{eq:formMu}
\mu_{t+1}={\eta\over \sqrt{t}} \erf\left({1-y^i\zeta_{t+1}}\over {\sigma\nu_t}\right)
\ee

\be
\label{eq:formNu}
\nu_{t+1}=\sqrt{\gamma_{t+1}^2\nu_t^2 + 2\gamma_{t+1}\mu_{t+1} y^i\zeta_{t+1} + \mu_{t+1}^2 K_{ii}}
\ee

\begin{algorithm}[H]
\SetLine
\KwData{Kernel function $\kappa$, training set $\cal S$, learning rate $\eta_0$, accuracy $\epsilon$}
\KwResult{$\alv$}
\caption{KEN-GURU($\kappa$,$\cal S$,$\eta_0$,$\epsilon$)}
\comm{initializations}
\ForAll {$m,n=1..m$} {
	$K_{mn} = \kappa(\xx^m,\xx^n)$
}
$\alv^0 \leftarrow \boldsymbol{0}$\;
$\nu_0 \leftarrow 0$\; 
$t \leftarrow 0$\;
\While{$\Delta L \geq\epsilon$} {
	\comm{randomize a sample}
	$i \leftarrow rand(M)$\;

	\BlankLine

	\comm{evaluate coefficients}
	Compute $\zeta_{t+1}$ (\eqref{eq:compZeta})\;
	Compute $\gamma_{t+1}$ (\eqref{eq:formGamma})\;
	Compute $\mu_{t+1}$ (\eqref{eq:formMu})\;

	\BlankLine
	\comm{update alphas}
	$\alv_{t+1}\leftarrow\gamma_{t+1}\alv_t$\;
	$\alpha_{t+1}^i\leftarrow \alpha_{t+1}^i + \mu_{t+1}$\;

	\BlankLine
	$t \leftarrow t + 1$\;
}
\Return $\alv$\;
\end{algorithm}

The correctness of the algorithm stems directly from that of GURU.

\section{Experiments}

In this section we present experimental results regarding the performance of KEN-GURU. We show how $\sigma$ affects the learned classifier and then compare KEN-GURU to SVM on USPS pairs and on the Ionosphere database (see \tabref{tab:binSizes} for details). For the USPS tasks, a polynomial kernel of degree $2$ was used and for Ionosphere, RBF with $\gamma=1$. The results are summarized in \tabref{tab:kerResults}.

\begin{figure}[btp]
  \centering
    \includegraphics[scale=0.4]{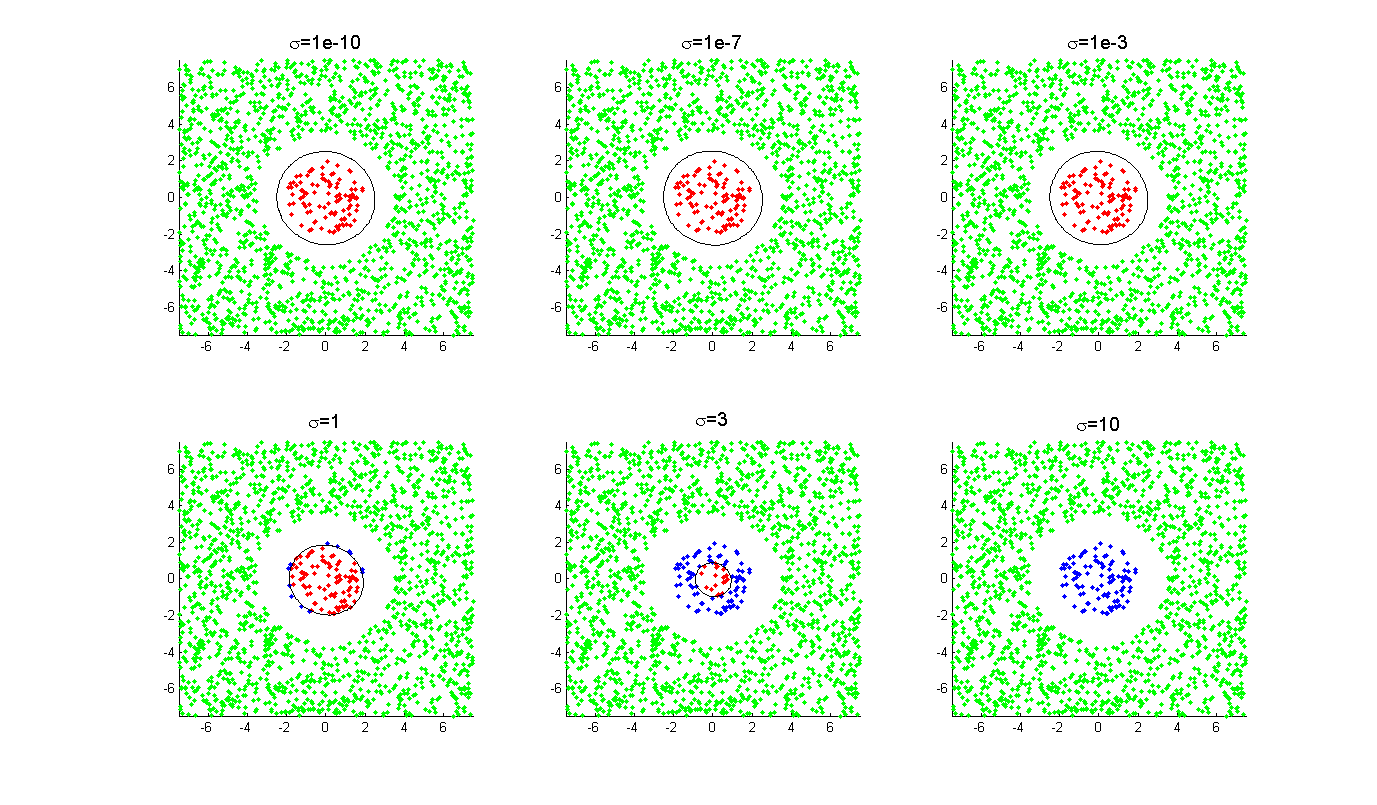}
  \caption{KEN-GURU performance on a radial data set. The green and red points indicate data points that were correctly classified (each color stands for one of the classes). Blue points indicate misclassification. The parameter $\sigma$ determines how distant is the effect of each data point. Note that for small values of $\sigma$, the behavior of the classifier is determined locally by the samples. For rather big $\sigma$, the effect is global, in the sense that the behavior of the classifier is determined by close as well as distant data samples.}
	\label{fig:radialKEN}
\end{figure}

Consider \figref{fig:radialKEN}, in which KEN-GURU classifiers trained for various values of the parameter $\sigma$ with a polynomial kernel of degree $2$ are presented. The toy probelm was synthesized by first generating uniformly points on $[-7.5,7.5]\times[-7.5,7.5]$. Points which fall within the ball of radius $2$ around the origin were assigned a positive label. Points which are more distant from the origin than $3.5$ units were taken as negative examples. Points which fell in between were dropped. Observe that increasing $\sigma$ puts extra emphasis on the number of samples in each class. Specifically, in the problem at hand, there are much more points outside the circle than inside. When $\sigma$ is rather small, the training is 'local' in the sense that each sample governs what happens in its immediate environment. On the contrary, when $\sigma$ is relatively big, the emphasis is on global tendencies.

\begin{table}
	\begin{center}
	    \begin{tabular}{ p{1.2cm} p{1.7cm} p{1.7cm} }
	    \hline
	    Name & GURU(\%) & SVM(\%) \\ \hline
		 Ionosp-here & 83.55 & 81.58 \\
		 diabetes & 68.59 & 66.67 \\
		 splice 1 vs. 2 & 92.28 & 92.28 \\ 	    
		 USPS 3 vs. 5 & 97.86 & 98 \\
	    USPS 5 vs. 8 & 98.29 & 98.71 \\
	    USPS 7 vs. 9 & 98.43 & 97.86 \\ \hline
	    \end{tabular}
		\caption{Results summary for KEN-GURU.}
		\label{tab:kerResults}
	\end{center}
\end{table}

On the Ionosphere databse, KEN-GURU performs significantly better than SVM. Recall that the outperformance of GURU on SVM in this case is consistent with the performance in the case of a linear kernel. This behavior is explained by the noisy nature of the Ionosphere database. For the USPS couples, KEN-GURU's performance is pretty similar to that of SVM.

\chapter{The Multiclass Case}
\label{section-multi}
In the previous chapters we have developed the binary algorhtm GURU, and its kernelized version KEN-GURU. In this chapter we will analyze anotther extension of the algorithm, for the case of multiclass cases.

The ideas that were presented in \chapref{section-GURU} may be generalized for the multi-class case. To that end, we first should generalize the loss function we are working with. This goal is acheived by solving the generalized problem of the adversarial choice. After establishing this reuslt we devise the effective robust loss function, and devise an optimization algorithm for it.

We relax the problem twice in order to solve it. First, we work with the sum-of-hinges loss function (\cite{Weston}). In addition, we use a superset of noise distribution, that contains all covariance matrix with a bounded maximal eigenvalue. By the end of the chapter we will prove that for the binary case the maximal eigenvalue and trace constraint give the same result.

The setting we address in the followings is of data drawn from $\mathcal{X}=\mathbb{R}^d$, accompanied by labels drawn from $\mathcal{Y}=\{1,2,\ldots,C\}$. The learning task is to train the weight vectors $\ww_1,\ww_2,\ldots,\ww_C$. The target classifier is $\phi:\mathcal{X}\to\mathcal{Y}$, defined by
\be
\phi(\xx;\ww_1,\ww_2,\ldots,\ww_C)=\max_{y\in\mathcal{Y}}\left[\ww_y^T\xx\right]
\ee

\section{Problem formulation}
In this section we formally describe the generalization of the learning task from the binary to the multiclass case. We show that the generalization culminates in a loss function which is the sum of several appropriate binary losses.

In \chapref{section-GURU} we have started our derivation from the hinge loss
\be
\ell_{hinge}(\xx^m,y^m;\ww) = [1-y^m\ww^T\xx]_+
\ee
The most common generalization of the hinge loss to the multi-class case is
\be
\ell_{mult}(\xx^m,y^m;\ww_1,\ldots,\ww_C)=\max_{y}\left[ \ww_y^T\xx^m - \ww_{y^m}^T\xx^m + \delta_{y,y^m} \right]
\ee
However, this loss function is not applicable in our framework (see \appref{app:mhinge}). Instead, we suggest to minimize the following surrogate loss function (Weston \& Watkins, e.g. ref):
\be
\ell_{sum}(\xx^m,y^m;\ww_1,\ww_2,\ldots,\ww_C)=\sum_{i\ne y}{\lf[1-(\ww_{y^m}-\ww_i)^T\xx^m\ri]_+}
\ee
which is a surrogate to the zero-one loss.

Let us write down the formulation of the problem in this case:
\be
\label{eq:multTask}
\min_{\ww_1,\ww_2,...,\ww_C}{
	\sum_{m}{
		\max_{\Sigma\in\Gamma_\beta}{
			\mathbb{E}_{\nn\sim\mathcal{N}(\bf{0},\Sigma)}{	
				\sum_{y'\neq y^m}{
						\left[1-(\ww_{y^m}-\ww_{y'})^T(\xx^m+\nn)\right]_{+}
				}
			}
		}
	}
}
\ee

where
\be\Gamma_\beta=\{\Sigma\in\textbf{PSD}\Big|\rho(\Sigma)\leq\beta\}\ee
and $\rho$ is the spectral norm of a matrix, defined by
\be
\rho(A)=\sqrt{\lambda_{\max}{(A^*A)}} \nonumber
\ee
Using this set we constrain the maximal power of noise that the adversary may spread in each primary direction.

\section{The adversarial choice}
In the followings we will focus on deriving the adversarial choise for the problem at hand. It appears that in the current setup, the solution is simpler than the one we had in \chapref{section-GURU}.

\subsection{Applying a spectral norm constraint}
Let us investigate what is the adversary's optimal way for spsreading the noise. The ideas of the development are similar to that of \thmref{thm:bacThm}.

Denote
\be
\Delta\WW_{y,y'}=\ww_y-\ww_y'
\ee
Using the same procedure we have employed in the binary case (see \secref{guru-loss-structure} and \eqref{eq:lossAfterInt} thereby) we can write \eqref{eq:multTask} as:
\be
\min_{\ww_1,\ww_2,...,\ww_C}{
	\sum_{m}{
		\max_{\Sigma\in\Gamma_\beta}{
			\sum_{y'\neq y^m}{
			L\left(\xx^m,+1;\Delta\WW_{y^m,y'},\Delta\WW_{y^m,y'}^T\Sigma\Delta\WW_{y^m,y'}\right)
			}
		}
	}
}
\ee
i.e. the task at hand is to optimize the effective loss function
\be
\label{eq:multiloss}
\ell_{sum}^{rob}(\xx^m,y^m;\ww_1,\ww_2,\ldots\ww_C,\beta)=\max_{\Sigma\in\Gamma_\beta}{
			\sum_{y'\neq 
		y^m}{L\left(\xx^m,+1;\Delta\WW_{y^m,y'},\Delta\WW_{y^m,y'}^T\Sigma\Delta\WW_{y^m,y'}\right)
			}
		}
\ee
Observe that in every appearance of $\lrh$, the label $y^m$ was replaced with $+1$. The reason for this change is that we are classifying using the weight vector $\ww_{y^m}-\ww_{y'}$. That is, our prediction is
\bea
{\left(\ww_{y^m}-\ww_{y'}\right)}^T\xx^m = \ww_{y^m}^T\xx^m-\ww_{y'}^T\xx^m
\eea
Our objective is, of course, to have $\ww_{y^m}^T\xx^m>\ww_{y'}^T\xx^m$, which corresponds to the label $+1$.
The next theorem specifies the adversarial choice of the covariance matrix $\Sigma$, and is the multi-class analog of \thmref{thm:bacThm}:
\thm
\label{thm:mbacThm}
The optimal $\Sigma$ in \eqref{eq:multiloss} is given by $\Sigma^*=\beta I$.
\ethm

\prf
In \lemref{lem:mono} we have shown that $L$ is monotone increasing in its $4^{th}$ argument.\newline
By the Cauchy-Schwartz inequality we have that
\be
\Delta\WW_{y^m,y'}^T\Sigma\Delta\WW_{y^m,y'}\leq\beta\|\Delta\WW_{y^m,y'}\|^2
\ee
On the other hand, it holds that for all $y'$
\be
\Delta\WW_{y^m,y'}^T\beta I \Delta\WW_{y^m,y'} = \beta\|\Delta\WW_{y^m,y'}\|^2
\ee
hence this upper bound is attained for all $C-1$ summands cuncurrenlty with  $\Sigma=\beta I$.\newline
The geometric interpertation of this result is that under the spectral norm constraint, the adversary will choose to spread the noise in an isothropic fashion around the sample point.
\eprf\newline

We thus get the following optimization problem:
\be
\min_{\ww_1,\ww_2,...,\ww_C}{
	\sum_{m}{
		\sum_{y'\neq y^m}{
			L(\xx^m,+1;\Delta\WW_{y^m,y'},\beta\|\Delta\WW_{y^m,y'}\|^2)
		}
	}
}
\ee
Applying the same terminology used in the binary case, we have:
\be
\min_{\ww_1,\ww_2,...,\ww_C}{
	\sum_{m}{
		\sum_{y'\neq y^m}{
			\ell_{hinge}^{rob}(\xx^m,+1;\Delta\WW_{y^m,y'},\beta)
		}
	}
}
\ee
and \eqref{eq:multiloss} equals
\be
\ell_{sum}^{rob}(\xx^m,y^m;w_1,w_2,...,w_C,\beta)=\sum_{y'\neq y^m}{\ell_{hinge}^{rob}(\xx^m,+1;\Delta\WW_{y^m,y'}),\beta}
\ee

\subsection{The connection to the trace constraint}
It is interesting to examine the reduction of the multiclass loss we have derived, to the binary case. Note that since we have used a substantially larger matrix collection, there is no apriori reason to expect that the results will coincide.

Taking $C=2$ brings us back to the binary case. We use $\ww_{+1}$, $\ww_{-1}$ for the weight vectors of the classes. By expanding \eqref{eq:multiloss}, we get

\be
\ell_{sum}^{rob}(\xx^m,y^m;\ww_{+1},\ww_{-1},\beta)=\ell_{hinge}^{rob}(\xx^m,+1;\ww_{y^m}-\ww_{-y^m},\beta)
\ee
If we take $\ww=\ww_{+1}-\ww_{-1}$, we end up with
\be
\ell_{sum}^{rob}(\xx^m,y^m;w_{+1},w_{-1},\beta)=\ell_{hinge}^{rob}(\xx^m,y^m;\ww,\beta)
\ee

It is interesting to observe that the resulting loss functions are identical, even though the constraints we put on the convariance matrices are different. In order to explain this phenomenon, let us go back the geometric intuition that we have given prior to the proof of \thmref{thm:bacThm}.\newline

\begin{figure}[h!]
  \centering
    \includegraphics[scale=0.5]{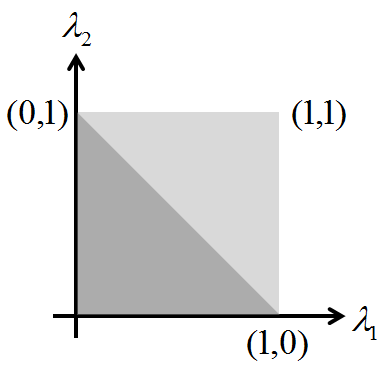}
  \caption{Visualization of $\Lambda_1$ and $\Gamma_1$ in the $2$-dimensional case. The axes represent the eigenvalues of $\Sigma$. The dark shaded region contains all the matrices having $\lambda_1+\lambda_2\leq 1$, i.e. corresponds to $\Lambda_1$. The light area corresponds to $\Gamma_1$, and consists of all the matrices with $\max{\{\lambda_1,\lambda_2\}}\leq 1$.}
	\label{fig:different_zones}
\end{figure}

Consider \figref{fig:different_zones}, which presents a visualization of $\Lambda_\beta$ and $\Gamma_\beta$ in the $2$-dimensional case. What we have shown in \thmref{thm:mbacThm}, is that the multiclass adversary will choose the point $(1,1)$. Under the trace constraint, however, the adversary will have to choose either $(1,0)$, $(0,1)$, or any other point lying on the line connecting them. Our geometric intuition says that all the power that was not spread perpendicularly to the separating hyperplane is irrelevant. Thus, when the adversary has to choose a directional noise, he would take the perpendicular direction. On the other hand, if we limit his action axis-wise (and not overall), he will surely choose to spread the noise equally over all of the axes.

\section{M-GURU: a primal algorithm for the multiclass case}
In the following we generalize GURU (that was presented in \secref{sec:guru_alg}) for the multiclass case. As a direct corrolary of the results presnted in previous chapters, we have that our loss function in this case is strictly-convex. Thus, we turn to devise an SGD procedure.

We shall begin by computing the gradient of $\ell_{sum}^{rob}(\xx^m,y^m;\ww_1,\ww_2,\ldots,\ww_C,\beta)$. For convenience, we write it in terms of the binary loss function $\lrh$:
\bea
  \nabla_{\ww_r}\ell_{sum}^{rob}(\xx^m,y^m&;&\ww_1,\ww_2,\ldots,\ww_C,\beta) = \nonumber \\
	&\mbox{}&\begin{cases}
   	 \sum_{y'\neq r}{\nabla_{\ww}\lrh(\xx^n,+1;\ww,\beta)\Big
			 |_{\ww=\ww_{y^m}-\ww_{y'}}} & \text{if } r=y^m\\
     -\nabla_{\ww}\lrh(\xx^n,+1;\ww,\beta)\Big |_{\ww=\ww_{y^m}-\ww_{r}}& \text{otherwise}\\
\end{cases}
\eea

Following the considerations that we have introduced in \secref{sec:guru_alg}, we devise an SGD procedure for the minimization task:

\begin{algorithm}[H]
\SetLine
\KwData{Training set $\cal S$, learning rate $\eta_0$, accuracy $\epsilon$}
\KwResult{$\ww$}
\caption{M-GURU($\cal S$,$\eta_0$,$\epsilon$)}
$\ww \leftarrow \boldsymbol{0}$\;
\While{$\Delta L \geq\epsilon$} {
	$m \leftarrow rand(M)$\;
	\For{$y'\in\{1,2,\ldots,C\}$} {
		$\ww_{y'} \leftarrow \ww_{y'}-{\eta_0\over \sqrt{t}}\nabla_{\ww_{y'}}\ell_{sum}^{rob}(\xx^m,y^m;\ww_1,\ww_2,...,\ww_C,\beta)$\;
	}
}
\Return $\ww$\;
\label{alg:mguru_basic}
\end{algorithm}

In \algref{alg:mguru_basic}, the notion of stochastic gradient was applied once, to the extent that our updates depend on a single sample in each iteration. It may be applied again, however. Instead of updating all the weight vectors concurrently, one might randomize which vector to update, as well. The resulting algrithm is

\begin{algorithm}[H]
\SetLine
\KwData{Training set $\cal S$, learning rate $\eta_0$, accuracy $\epsilon$}
\KwResult{$\ww$}
\caption{M-GURU-$S^2$($\cal S$,$\eta_0$,$\epsilon$)}
$\ww \leftarrow \boldsymbol{0}$\;
\While{$\Delta L \geq\epsilon$} {
	$m \leftarrow rand(M)$\;
	$y'\leftarrow rand(C)$ {
	$\ww_{y'} \leftarrow \ww_{y'}-{\eta_0\over \sqrt{t}}\nabla_{\ww_{y'}}\ell_{sum}^{rob}(\xx^m,y^m;\ww_1,\ww_2,...,\ww_C,\beta)$\;
	}
}
\Return $\ww$\;
\label{alg:mguru_sg_sq}
\end{algorithm}

\section{Experiments}
\begin{figure}[h!]
  \centering
    \includegraphics[scale=0.6]{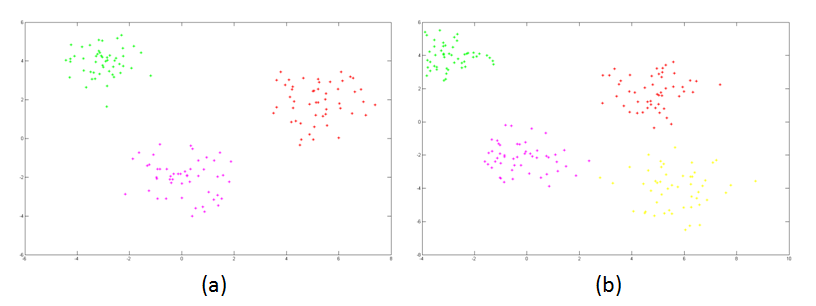}
  \caption{The toy problems used in the testing of M-GURU and M-GURU-$S^2$. (a) Toy-3. (b) Toy-4.}
	\label{fig:mults}
\end{figure}

M-GURU and M-GURU-$S^2$ were tested on toy problems, USPS and a couple of UCI databases (\cite{UCI}). The datasets are detailed in \tabref{tab:multSizes}. In Toy-3 and Toy-4 each class is a Gaussian distribution. These problems are visualized in \figref{fig:mults}. The rsults are summarized in \tabref{tab:multResults}.

\begin{table}
	\begin{center}
	    \begin{tabular}{  p{1.2cm} p{1.5cm}  p{1.5cm}  p{1.5cm}  p{1.5cm} p{1.5cm} }
	    \hline
	    Name & \#Training samples & \#Cross-validation samples & \#Test samples & \#features & \#classes\\ \hline
	    Toy-3 & 200 & 200 & 200 & 2 & 3 \\
	    Toy-4 & 200 & 200 & 200 & 2 & 4 \\
	    USPS 3,5,8 & 1200 & 1050 & 1050 & 256 & 3 \\
	    USPS 0-9 & 3000 & 2000 & 6000 & 256 & 10 \\
		 splice & 1000 & 1000 & 1190 & 60 & 3 \\
		 wine	& 50 & 50 & 78 & 13 & 3 \\ \hline
	    \end{tabular}
		\caption{Description of the databases used in the binary case}
		\label{tab:multSizes}
	\end{center}
\end{table}

Observe that the performance of M-GURU is similar to that of SVM. Nontheless, it should be noted that SVM slightly outperforms M-GURU. This difference is explained by the fact that M-GURU is based on the sum-of-hinges loss function, which is a looser surrogate of the zero-one loss than the SVM multi-hinge loss function. We have tested the relative performance of M-GURU and M-GURU-$S^2$ on the toy-3 dataset.

\begin{figure}[h!]
  \centering
    \includegraphics[scale=0.6]{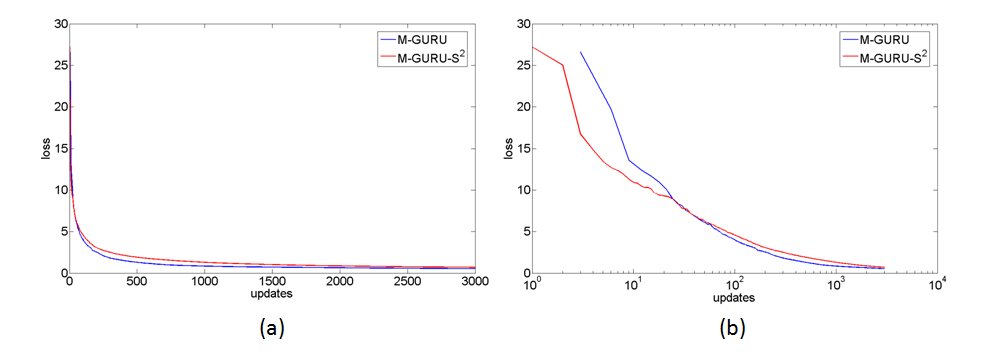}
  \caption{A typical run of M-GURU and M-GURU-$S^2$ on the toy-3 dataset. The loss is plotted against the number of updates that were performed. The $S^2$ variant appears to have an advantage in the descent phase. In the convergence phase, however, M-GURU takes the lead. Overall, the performance of both variants is pretty similiar. (a) linear scale. (b) semi-logarithmic scale.}
	\label{fig:variants}
\end{figure}

\begin{table}
	\begin{center}
	    \begin{tabular}{ p{1.2cm} p{1.7cm} p{1.7cm} p{1.7cm} }
	    \hline
	    Name & M-GURU(\%) & M-GURU-$S^2$(\%) & SVM(\%) \\ \hline
	    Toy-3 & 98.67 & 98 & 98.67\\
	    Toy-4 & 96 & 96 & 96\\
	    USPS 3,5,8 & 94.67 & 94.57 & 94.857\\
	    USPS 0-9 & 92.78 & 92.7 & 92.85 \\
	    splice & 89.08 & 89.08 & 89.5 \\
	    wine & 92.31 & 91.03 & 92.31\\ \hline
	    \end{tabular}
		\caption{Summary of the results.}
		\label{tab:multResults}
	\end{center}
\end{table}

We observe that M-GURU outperforms the $S^2$ variant. Our experiments show that the empirical behavior of the classifiers stabilizes a significant time before the optimization process converges. Thus, M-GURU-$S^2$ may be used to learn classifiers more quickly.

\chapter{Discussion}
\label{section-discussion}

\section{Contribution}
In this work we presented a new robust learning framework. In our framework we minimize the expected loss over a spreading of the sample points. Each displacement is assumed to take place with a probability that depends on its distance from the original point. Thus, we effectively replace each point with a fading cloud.

We have analyzed the case of Gaussian noise distribution, where the underlying loss measure is the hinge-loss. In this case, we have shown that the resulting effective loss function is a smooth strictly-convex upper-approximation of the hinge-loss, denoted $\lrh$. One of the main advantages of this loss function, is its parameter $\sigma$ that has a clear meaning: the variance of the noise that contaminates the data. Similarly to SVM, our algorithm, named GURU, depends on a single parameter. A significant difference is the ability to
assign a value to this parameter. In the case of SVM, for a long time all that was known on this parameter is that it controls the tradeoff between the training error and the margin of the classifier. \cite{XuCaMa09} have shown that SVM is equivalent to a robust formulation in which the parameter corrsponds to the radius of a rigid ball in which the sample point may be displaced. This result, however, relates the parameter with the entire data set. Thus, it is
still difficult to tune it. In our method, $\sigma$ is the magnitude of noise that possibly corrupts each sample point, hence it might be evaluated from physical consideration, such as the process that generates the data, etc. Without putting extra effort, we are able to point out an alternative explanation for non-regularized SVM's lack of ability to generalize. We have shown that as $\sigma$ tends
to $0$, $\lrh$ coincides asymptotically with the hinge loss. Thus, non-regularized SVM may be understood as not trying to acheive robustness to perturbations, hence it tends to overfit the data.
We have shown that $\lrh$ may be written as a perspective of a smooth loss function (denoted $f$), where the scaling factor is $\sigma\|\ww\|$. This representation suggests that the robust framework we have developed introduces a multiplicative regularization. Using both this
representation we have derived a dual problem. The dual formulation depends on the actual loss function $f$ only via its conjugate dual. Thus, it is possible to plug into the same formulation some other losses that follow certain conditions. In particular, as we have demonstrated in \chapref{section-dual}, there is a tight connection between approximations of the loss function and relaxations of the dual problem. We believe that applying the same technique we have
apllied here to other loss functions will result in new robust learning algorithms. The connection between the primal loss and the resulting dual shold be investigated more throughly.
The algorithmic approach we have taken in this work is rather simplistic. Due to the fact that our objective is strictly-convex, many off-the-shelf convex optimization algorithms may be used. Our method of choice was stochastic gradient descent. Furthrmore, if there is a bound on the norm of the optimal classifier (as in SVM. see \cite{Pegasos} for details), it is probably possible to use it in order to achieve even faster algorithms. Specifically, subject to such a bound, we may restrict the optimization problem to a ball around the origin. In this ball, it is possible that our loss function is strongly-convex,hence it can be optimized using more aggressive procedure (\cite{Sha08}). Our generalization to Mercer kernels, is done based on the primal formulation. In
order to compute the updates fast ($O(M)$), we have shown how to maintain the value of the norm of the classifier in $O(1)$ based on pre-computed values. This technique may be employed in Pegasos, e.g, in order to perform the projection step efficiently.

\section{Generalizations}
The framework we have introduced may be generalized in couple of interesting directions. Obviously, various families of noise distributions may be plugged into the model. One particularly interesting is the class of all probabilty distributions having a specific first and second moment. \cite{Vanden} have shown that the probability of a set defined by quadratic inequalities may be computed using semidefinite programming. In addition, they have shown that the optimum is acheived over a discrete probability distribution. We conjecture that a similar technique may be employed in our case, in order to show that the optimum of the loss expectation is attained over a discrete distribution. In addition, the same framework can be used in order to explore more convex perturbations. For example, in the field of computer vision it is possible to assume that the adversary rotates or translates the sample, and that the distribution of these perturbations is chosen adversely. In order to make this practical, it is crucial to understand in which cases the integration and integration of the loss are possible.

Regarding the theoretical aspects of this work, it still remains to show how to derive performance bounds for the introduced framework. In particular, it is interesting to understand what kind of gurantees can be derived for the general perspective-optimization framework we have discussed.

\bibliographystyle{plain}
\bibliography{rob_bib}

\begin{appendices}
\chapter{Single-Point Algorithms}
\label{app:singAlgs}
The object of this work is to learn classifiers that are robust to noise. As discussed, a possible way to achieve this goal is by applying an adversarial framework. The most important issue in this case is designing an effective adversary. While in the previous chapters of the work we explored more sophisticated adversaries, it is nice to end the journey with a rather simple mathematical formulation. The binary version of the algorithms was extensively studied. We review the result here for the sake of a complete presentation. A simple generalization for the multiclass case is presented subsequently.

\section{Problem presentation}

Maybe the simplest action that the adversary can take at test-time is displacing a test point, in such a way that will cause this point to be missclassified. If we limit the freedom given to the adversary, it might not be able to corrupt the classification of the point, but rather only reduce the associated confiedence. The model that we will explore in the followings grants the adversary the ability to displace a sample point within a ball centered at the original point. 
\newline
In order for the learned classifier to be robust to such displacements, we should modify the objective of the learning task. In the following we present and anlyze one way to do it, by optimizaing the worst-case scenario:
\be
\label{eq:ASVCunDec}
\min_{\ww}{\max_{\|\Delta\xx^m\|\leq\delta:\;m=1..M}{
	{\lambda\over{2}}\|\ww\|^2 +
	\sum_{m=1}^M{
		{
			\left[1-y^m\ww^T(\xx^m+\Delta\xx^m)\right]_+
		}
	}
}
}
\ee
This formulation has an additive structure, in which each term $\Delta\xx^m$ appears exactly once. We use these properties in order to decouple the optimization problem. The learning task at hand in this case is thus
\be
\label{eq:ASVCdec}
\min_{\ww}{
	{\lambda\over{2}}\|\ww\|^2 +
	\sum_{m=1}^M{
		\max_{\|\Delta\xx^m\|\leq\delta}{
			\left[1-y^m\ww^T(\xx^m+\Delta\xx^m)\right]_+
		}
	}
}
\ee

Recall that in the general SVM setting, one tries to minimize the hinge loss: \be
\ell_{\text{hinge}}(\xx,y;\ww)=[1-y\ww^T\xx]_+
\ee

\eqref{eq:ASVCdec} can be interpreted as optimizing the effective loss function
\be
\label{eq:hingerob}
\ell_{\text{hinge}}^{\text{rob}}(\xx,y;\ww)=
\max_{\|\Delta\xx\|\leq\delta}{[1-y\ww^T(\xx+\Delta\xx)]_+}
\ee
We say that this loss function is robust, in the sense that it represents the worst-case loss subject to the potential action of the adversary.

\section{Computing the optimal displacement}

\begin{figure}[h!]
  \centering
    \includegraphics[scale=1]{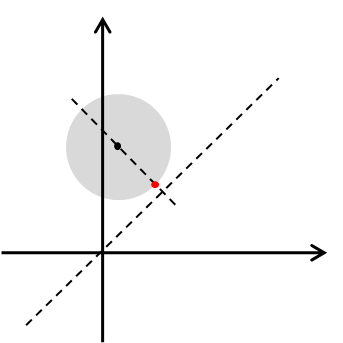}
  \caption{The adversarial displacement employed by ASVC}
	\label{fig:ASVCdep}
\end{figure}

In order to derive a closed form for the loss function $\ell_{\text{hinge}}^{\text{rob}}$, we should explore the nature of the adversarial choice in our model. Intuitively, the adversary will try to relocate the point to the wrong side of the seperating hyperplane. For this end, it is pointless to move the point along any axes not orthogonal to the seperating hyperplane. This idea is visualized in \figref{fig:ASVCdep}. We will now prove this simple theorem:

\thm
\label{thm:ASVCthm}
The optimum of the maximization in \eqref{eq:hingerob} is acheived at $\xx_{\text{opt}} = \xx-\delta{\ww\over{\|\ww\|}}$
\ethm

\prf
First we observe that the function $f(z) = [1-z]_+$ is a monotone non-increasing function of its argument $z$. Thus, maximizing $f(z)$ is equivalent to minimizing $z$. By the Cauchy-Schwartz inequality, we have that $|y\ww^T\Delta\xx|\leq\|\ww\|\cdot\|\Delta\xx\|$, with equality iff $\Delta\xx$ is proportional to $\ww$. Therefore, the minimal value possible is attained at $\Delta\xx_{\text{opt}} = -\delta{\ww\over{\|\ww\|}}$. We conclude that $\xx_{\text{opt}} = \xx-\delta{\ww\over{\|\ww\|}}$ as claimed.
\eprf

Plugging the result of the theorem above into \eqref{eq:hingerob} we end up with
\be
\label{eq:hingerobFinal}
\ell_{\text{hinge}}^{\text{rob}}(\xx,y;\ww)=
{[1-y\ww^T\xx+\delta\|\ww\|]_+}
\ee

\section{ASVC: Adversarial Support Vector Classification}
The fact that \eqref{eq:hingerob} has a simple closed-form solution allows us to employ the algorithmic scheme of alternating optimization for \eqref{eq:ASVCunDec}. The structure of the algorithm is quite simple: \begin{enumerate}
\item Alternately:
	\begin{enumerate}
		\item Optimize for $\ww$ \label{ASVC_SVM}
		\item Optimize for $\Delta\xx^1$, $\Delta\xx^2$,..., $\Delta\xx^M$ \label{ASVC_closed}
	\end{enumerate}
Until convergence.
\end{enumerate}

Notice that \ref{ASVC_SVM} is nothing more than an SVM taking the displaced points as input. Furthermore, \ref{ASVC_closed} has a closed-form solution as we have proved in \thmref{thm:ASVCthm}. Thus, to solve for the optimal classifier, any off-the-shelf SVM solver can be used. We end up with \algref{alg:ASVC}.

\begin{algorithm}[H]
\SetLine
\KwData{Training set $\cal S$, radius $\delta$, tradeoff $\lambda$}
\KwResult{The weight vector $\ww$}
\caption{ASVC($\cal S$, $\delta$, $\lambda$,$T$,$k$)}
$\ww \leftarrow \boldsymbol{0}$\;
\Repeat {\text{convergence}}{
	$\Delta\xx^m \leftarrow -\delta{\ww\over{\|\ww\|}}$\;
	$\tilde{\cal S}\leftarrow \{\xx^m+\Delta\xx^m\}_{\xx^m\in\cal S}$\;
	$\ww \leftarrow \text{solveSVM}(\tilde{\cal S},\lambda)$
}
\Return $\ww$\;
\label{alg:ASVC}
\end{algorithm}

\section{The Multiclass Case}
Pretty similar ideas can be adopted in order to generalize ASVC for the multiclass case.

The multi-hinge loss is defined as
\be
\ell_{mult}(\xx^m,y^m;\ww_1,\ww_2,...,\ww_C) =
\max_{y=1,2,...,C}{\left[\delta_{y,y^m}-(\ww_{y^m}-\ww_y)^T\xx^m\right]}
\ee

Using the notions of the previous section, we define

\be
\ell_{mult}^{\text{single}}(\xx^m,y^m;\ww_1,\ww_2,...,\ww_C) =
\max_{\|\Delta\xx\|\leq\delta}
	{\max_{y=1,2,...,C}
		{\left[\delta_{y,y^m}-(\ww_{y^m}-\ww_y)^T(\xx^m+\Delta\xx)\right]}}
\ee

Note the order of maximization can be changes, i.e.
\be
\ell_{mult}^{\text{single}}(\xx^m,y^m;\ww_1,\ww_2,...,\ww_C) =
\max_{y=1,2,...,C}
	{\max_{\|\Delta\xx\|\leq\delta}
		{\left[\delta_{y,y^m}-(\ww_{y^m}-\ww_y)^T(\xx^m+\Delta\xx)\right]}}
\ee

Applying a slight variation of \thmref{thm:ASVCthm}, we conclude with
\bea
\ell_{mult}^{\text{single}}(\xx^m,y^m&;&\ww_1,\ww_2,...,\ww_C) \nonumber \\
&=&\max_{y=1,2,...,C} { 
	{\left[\delta_{y,y^m}-(\ww_{y^m}-\ww_y)^T
		\left(\xx^m
			-\delta{{\ww_{y^m}-\ww_y}\over{\|\ww_{y^m}-\ww_y\|}}
		\right)
	\right]}
} \nonumber \\
&=& \max_{y=1,2,...,C} { 
	{\left[\delta_{y,y^m}-(\ww_{y^m}-\ww_y)^T\xx^m + \delta\|\ww_{y^m}-\ww_y\|
	\right]} }  
\eea

\section{Related work}
Our ASVC algorithm is a mirror reflection of TSVC presented in (Bi \& Zhang, NIPS04). TSVC performs alternating optimization, each time replacing the set of training samples with $\{\xx^i+y^i\delta^i{\ww\over\|\ww\|}\}$, which are more distant from the separator (thus, easier to classify). The idea there is to address the case in which noisy data distracts the classifier, by using the shifted training sets. \newline

\begin{figure}[h!]
  \centering
    \includegraphics[scale=1]{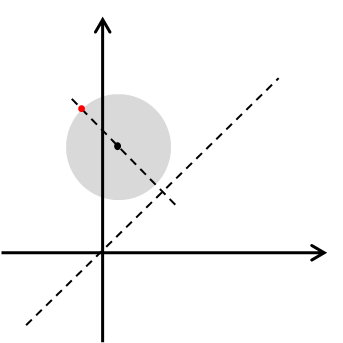}
  \caption{The displacement employed by TSVC}
	\label{fig:TSVCdep}
\end{figure}

\chapter{Diagonal Covariance}
\label{app:diagCov}
In this appendix we discuss the case in which the adversary is constrained to choose a diagonal covariance matrix. This setting corresponds to the case when the noise is alligned to the primary axes. In this case we are able to give a closed form analytical result, subject to a bounded trace constraint on the covariance matrix.

The adversarial choice problem can can be written
\be
\max_{\Sigma=\text{diag}(a_1,a_2,\ldots,a_d)\;tr(\Sigma)\leq\beta}{L(\xx^m,y^m;\ww,\ww^T\Sigma\ww)}
\ee
Let us expand
\bea
\ww^T\Sigma\ww &=& \ww^T\text{diag}(a_1,a_2,\ldots,a_d)\ww \nonumber\\
&=& \sum_{i=1}^d{a_i {w_i}^2}=\boldsymbol{a}^T\ww^{\cdot 2}
\eea
where $\ww^{\cdot 2}$ represents the coordinate-wise product of $\ww$ with itself. Let $i^*$ be the index of the maximal entry in $\ww^{\cdot 2}$. It hold that 
\be
\ww^T\Sigma\ww\leq \sum_i{\beta_i}w_{i^*}^2
\ee
Using the same argumentation as in \chapref{section-GURU}, we conclude that the adversary will choose the covariance matrix
\be
\Sigma^*=\beta\boldsymbol{e}_{i^*i^*}
\ee
where $e_{ij}$ is the matrix having zeros in all of its entries beside $(i,j)$, where it takes the value $1$. The geometric meaning of this result is that the adversary will choose to spread the noise in a single direction, along the primary axis that creates the biggest angle with the separating hyperplane.

\begin{figure}[h!]
  \centering
    \includegraphics[scale=0.7]{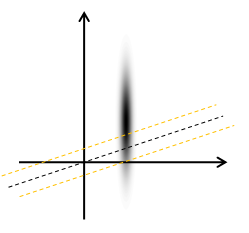}
  \caption{Under the diagonal covariance restriction, the adversary will choose to spread the noise in a unique direction. This direction is the one that creates the biggest angle with the separating hyperplane.}
	\label{fig:different_zones}
\end{figure}

\chapter{Using the Multi-Hinge Loss}
\label{app:mhinge}
The most common generalization of the hinge loss for the multiclass case is the following loss function
\be
\ell_{mult}(\xx^m,y^m;\ww_1,\ldots,\ww_C)=\max_{y}\left[ \ww_y^T\xx^m - \ww_{y^m}^T\xx^m + \delta_{y,y^m} \right]
\ee
(see \cite{Crammer02}). In this appendix we point out some of the issues that made us choose to work with the sum-of-hinges loss function and not with the one above.

If we plug the multi-hinge loss into our framework, we get the following learning problem:
\be
\min_{\ww} \sum_m \max_{\Sigma\in S}\int p(\hat\xx|\xx^m;\Sigma) \max_{y}\left[ \ww_y^T\hat\xx - \ww_{y^m}^T\hat\xx + \delta_{y,y^m} \right] d\hat\xx  
\ee
Define $\Delta\ww_{y,y^m} = \ww_y - \ww_{y^m}$ and write:
\be
\min_{\ww} \sum_m \max_{\Sigma\in S}\int p(\hat\xx|\xx^m;\Sigma) \max_{y}\left[ \Delta\ww_{y,y^m}\hat\xx + \delta_{y,y^m}\right]d\hat\xx
\ee
And for Gaussian noise this is:
\be
\min_{\ww} \sum_m \max_{\Sigma\in S}c|\Sigma|^{-0.5}\int e^{-{1\over 2}\nn^T\Sigma^{-1}\nn}\max_{y}\left[ \Delta\ww_{y,y^m}\xx^m+ \Delta\ww_{y,y^m}\nn+ \delta_{y,y^m}\right]d\nn
\ee
The ability to understand the solution of the adversarial choice problem in this case, is connected to the ability to understand the expectation of the maximum of a set of normal random variables. This problem probably does not have an analytical solution (see \cite{maxHard}).

\subsubsection{unidirectional noise}
In another approach we have studied, we assumed an adversary that spreads the noise in a single direction. The motivation for this kind of adversary is the solution to the adversarial choice problem in the binary case.

We formulate the problem by letting the adversary to choose a unit length vector. Thus, in the case of unidirectonal noise, the task that the adversary faces is:
\be
\max_{\vv:\|\vv\|\le 1}{\int_\mathbb{R}{\mathcal{N}_z(0,\sigma^2)\max_y\left[\Delta\ww_{y,y^m}^T\xx^m+\Delta\ww_{y,y^m}^T\nn+\delta_{y,y^m}\right]}}dz
\ee

The integrand (excluding the pdf) is a piecewise linear function. The knees of this function as well as the slopes of the linear sections are strongly dependent on $\vv$. Nontheless, it is impossible to find a closed form solution for the position of the knees. Therefore, we find this direction inapplicable in our case, as well.

\end{appendices}
\newpage

\end{document}